%% file: main.tex
\documentclass[]{fairmeta}

\usepackage{template/macro}
\input{packages}

\pdfpageattr{/Rotate 0}

\title{AutoformBench: A Lean Formalization Benchmark for Advanced Mathematics}
\author[*,1]{Charles Arnal}
\author[*,1,2]{Ahmad Rammal}
\author[\dagger,3]{Niket Patel}
\author[\dagger,2]{Fabian Gloeckle}
\author[\dagger,5]{Bartosz Piotrowski}
\author[2,4]{Amaury Hayat}
\author[\dagger,3]{Julia Kempe}
\author[1]{Rémi Munos}
\author[1]{Vivien Cabannes}
\affiliation[1]{FAIR at Meta}
\affiliation[2]{CERMICS, ENPC, Institut Polytechnique de Paris}
\affiliation[3]{New York University}
\affiliation[4]{Korea Institute for Advanced Study}
\affiliation[5]{Axiom Math}
\contribution[*]{Joint first authors (randomized order)}
\contribution[\dagger]{Work done while at Meta}

\abstract{%
AI systems are increasingly capable of producing machine-checked Lean proofs of advanced mathematical results. Real mathematical formalization, however, is not isolated theorem proving: it requires building the definitions and intermediate lemmas that connect a result to a formal library. Existing evaluations often provide isolated theorem statements in settings where much of this supporting theory is already available in Mathlib. We introduce AutoformBench, a benchmark of 341 Lean 4 formalization tasks drawn from 24 advanced mathematical sources. Lean-experienced mathematicians formalized and audited each target statement, leaving the corresponding Lean proof to the evaluated agent. Each problem is embedded in a broader mathematical development: agents receive the informal proof but must translate it into Lean while constructing the missing definitions, intermediate lemmas, and supporting infrastructure. We compare the performance of frontier models on these problems. We find a largely shared difficulty frontier: success depends less on mathematical domain than on models' ability to construct missing formal infrastructure.\\
Data and code at \url{https://github.com/facebookresearch/autoformbench}.
}

\date{\today}
\correspondence{\\Charles Arnal at \email{charlesarnal@meta.com}\\
Ahmad Rammal at \email{rammal@meta.com}}

\begin{document}

\maketitle
\raggedbottom

\begin{figure}[H]
  \centering
  \begin{minipage}[b]{0.46\textwidth}
    \centering
    \includegraphics[width=\linewidth]{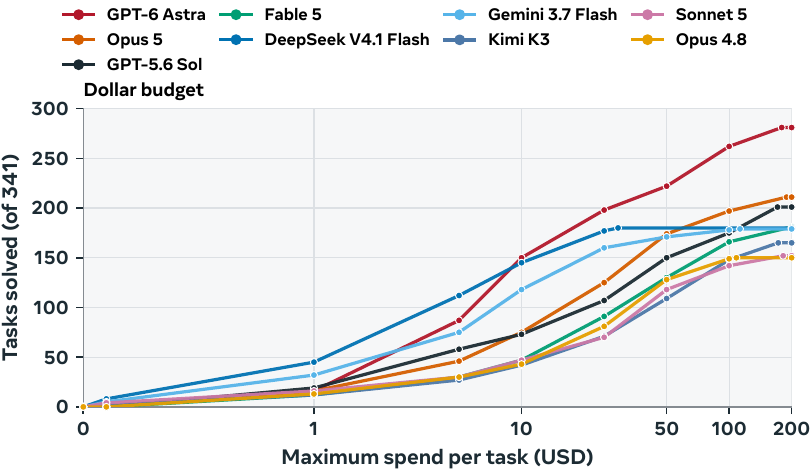}
  \end{minipage}\hspace{0.015\textwidth}%
  \begin{minipage}[b]{0.52\textwidth}
    \centering
    \includegraphics[width=\linewidth]{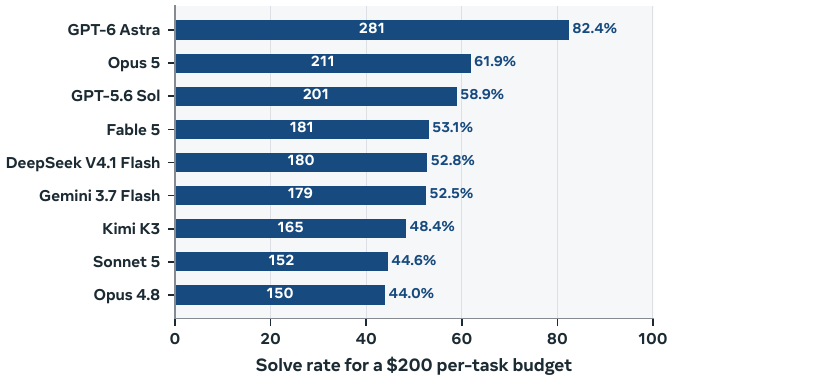}
  \end{minipage}
  \caption{Performance and cost-efficiency of frontier models on AutoformBench. Left: number of tasks completed under increasing per-task dollar budgets. Right: solve rate at a \$200 per-task budget.}
  \label{fig:main-results}
\end{figure}

\section{Introduction}

As language models produce increasingly sophisticated mathematical arguments,
verifying their output becomes a central bottleneck. Without reliable verification, scaling mathematical generation risks producing a growing body of plausible but incorrect results that is increasingly difficult to audit. Proof assistants, such as Lean, offer a
way to make verification mechanical. In Lean~4, a small kernel checks
every logical step, so an accepted proof provides a precise certificate relative
to its formal statement and assumptions~\citep{moura2021lean4}.

Producing that certificate for advanced mathematics is not merely a matter of
finding the final proof. Formalization is cumulative: before proving a theorem,
one must identify or construct the definitions, intermediate lemmas, and
connections to Mathlib on which it depends~\citep{mathlib2020}.

This cumulative work is underrepresented in current evaluations. Neural theorem
proving has progressed from language-model proof search
~\citep{polu2020generative,lample2022hypertree} to systems capable of solving
Olympiad and research-level problems
~\citep{hubert2026alphaproof,achim2025aristotle,chen2026fel}, but benchmarks such
as miniF2F, PutnamBench, and ProofNet primarily test self-contained problems
from mathematical competitions or undergraduate courses whose formal
vocabulary is already supplied
~\citep{zheng2022minif2f,tsoukalas2024putnambench,azerbayev2023proofnet}.
Success on these tasks does not tell us whether an agent can navigate a new
mathematical development, build its missing infrastructure, or do so
efficiently.

The Jacobian Challenge~\citep{buzzard2026jacobian} points toward this broader
regime: its compact statements sit atop definitions and results that demand
substantial formal development.  Our proposed benchmark, AutoformBench, asks how far current agents can
operate in this regime across mathematical domains. Each of its 341 problems
provides a human-audited Lean statement, its local formal context, and the
corresponding excerpt from one of 24 advanced mathematical sources.  The
agent must prove the target by interacting agentically with the provided material and with mathlib and by constructing any missing auxiliary definitions
and lemmas. This makes AutoformBench a challenging test of long-horizon repository-level formalization - even the most state-of-the-art models require dozens of millions of tokens to complete 50\% of the tasks (see Figure~\ref{fig:main-results}), highlighting the need for much more cost-efficient approaches before autoformalization can effectively scale to the mathematical literature and new research production.

\paragraph{Contributions.}
\begin{itemize}
  \item A benchmark of 341 audited Lean~4 proof-completion problems drawn from 24
        advanced mathematical sources across diverse fields.
    \item An agentic evaluation environment powered by state-of-the-art Lean tools for reliable and convenient model testing.
  \item A controlled evaluation of nine frontier models, together with analyses of efficiency,
        generated infrastructure, and the boundary between solved and unsolved
        problems.
\end{itemize}

The dataset and the evaluation code can be found at \url{https://github.com/facebookresearch/autoformbench}.

\section{Related Work}

\paragraph{Benchmarks for formal mathematics.}  miniF2F and
PutnamBench evaluate proofs
of fixed formal statements drawn from high-school and undergraduate
competitions~\citep{zheng2022minif2f,tsoukalas2024putnambench}, while
FormalProofBench pairs advanced undergraduate and graduate natural-language
problems with Lean statements and evaluates proof production in a tool-enabled
agent loop~\citep{ravi2026formalproofbench}.
ProofNet, Lean Workbook, FormL4, NuminaMath-LEAN, and FormalMATH scale the
construction or evaluation of paired informal and formal problems
~\citep{azerbayev2023proofnet,ying2024lean,lu2024process,numina2025lean,yu2025formalmath}.
These collections primarily consist of independent theorem-level items drawn
from competitions, educational material, or existing formal libraries. FATE
instead focuses on algebra at increasing levels of difficulty:
FATE-M contains textbook exercises, while FATE-X reaches beyond PhD qualifying
exams and Mathlib coverage~\citep{jiang2026fate}. Each file centers on one final
theorem with one proof hole; 38\% of FATE-X tasks additionally introduce new
definitions, averaging 2.4 among those tasks. SorryDB~\citep{letson2026sorrydb} moves toward realistic
library context by extracting open proof obligations from Lean repositories.
FormalQualBench contains 23 graduate-level theorem statements formalized in
Lean and verified by a Lean expert for evaluating practical formalization
workflows~\citep{formalqualbench2026}, while
Formal Conjectures provides an evolving collection of solved and open
research-level problems formalized in Lean~\citep{formalconjectures}.

\paragraph{Research-level autoformalization.}
Autoformalization studies the translation of informal
mathematics into formal language~\citep{wu2022autoformalization}. Recent work has begun to formalize sustained mathematical developments:
general topology in Megalodon~\citep{urban2026formal}, real- and convex-analysis
textbook material~\citep{wang2026m2f}, and a graduate textbook using a largely
autonomous multi-agent workflow~\citep{gloeckle2026automatic}. AutoformBot extends this approach
across 26 textbooks using a shared multi-agent pipeline
~\citep{rammal2026formalizing}. More recent systems study document-to-project
workflows for mathematical papers, long-horizon research developments, and mathematical discovery
~\citep{milikic2026leanflow,zhang2026leanmarathon,nie2026formatheoria,
moakhar2026theo, patel2026learningdiscoverinterestingmathematics}.
Other projects formalize landmark research results, including sphere packing in
dimension eight and De Giorgi--Nash--Moser theory
~\citep{hariharan2026spherepacking,armstrong2026formalizationgiorginashmosertheorylean}, and Anthropic's formalization of Fermat's Last Theorem~\citep{anthropicFLT2026}.
The Jacobian Challenge~\citep{buzzard2026jacobian} similarly highlights the
substantial infrastructure needed for a compact advanced mathematical
development.

AutoformBench
complements these efforts by evaluating proof completion across many developments
derived from advanced mathematical sources, with human-audited statements, access to the
informal source, and rigorous verification of candidate solutions. It is designed to measure a model's capability to interact with a body of mathematics, testing the skills required to formalize a large body of advanced mathematics - to our knowledge the first large-scale benchmark to do so.

\section{Benchmark Construction}
\label{sec:benchmark}
AutoformBench consists of challenging statements drawn from advanced mathematical sources spanning
diverse mathematical domains and formalized by human experts.

\paragraph{Source material.}
We selected $24$ openly available advanced mathematical sources from a variety of domains representative of the diversity of modern mathematics, ranging from algebraic geometry to the theory of probability. Most of these come from the excellent MIT OpenCourseWare selection of open courses~\citep{MITOCW}. The full list of selected sources and their provenance can be found in Table~\ref{tab:source-provenance} in the Appendix.

Within each book, we then selected statements with challenging proofs following a balanced distribution across chapters and subtopics, so as to maximize the diversity of the resulting tasks.
We deliberately selected material beyond direct
Mathlib~\citep{mathlib2020} coverage, so that success requires constructing missing formal infrastructure rather than retrieving an existing theorem, while retaining
enough nearby foundations for meaningful progress within a bounded agent run.
Appendix~\ref{app:mathlib-proximity} quantifies proximity to the pinned Mathlib revision.

\paragraph{Formalization protocol}
Human experts (at least PhD-level) from the annotation company Labelbox \citep{Labelbox} were tasked with formalizing the selected theorems, as well as all the definitions and lemmas needed to state them, in Lean~4~\citep{moura2021lean4} (version 4.32).
Each resulting formalization was then reviewed by an independent human expert, as well as by the authors assisted by frontier models.

\paragraph{Output}
This process resulted in 341 self-contained Lean workspaces containing 1{,}155
graded proof obligations. Each workspace has one headline target; one or more
proposition-valued proof bodies are replaced by the Lean keyword
\texttt{sorry}. At evaluation time, the tested model must replace every graded
\texttt{sorry} with a correct Lean proof and may introduce additional
definitions and helper lemmas. A task counts as solved only when the Lean kernel
accepts all of its graded obligations. We release all 341 tasks in Harbor
format. Appendix~\ref{app:benchmark-structure} gives the complete distribution
of obligations and characterizes the formal context visible in the target
statements.

\section{Evaluation Protocol}
\label{sec:evaluation-protocol}

\paragraph{Workspace and verification.}
For each problem, we construct a self-contained Lean workspace around the
audited target statement. It contains a frozen file containing the statement and its
supporting definitions, together with an editable copy of this file. The model must complete the statement's proof in the editable file and may add further definitions
and lemmas. A machine-readable specification identifies the results to grade
and the foundational axioms that may be used. The workspace also contains the
corresponding textbook statement, surrounding definitions, and informal proof. We execute these workspaces using Harbor~\citep{merrill2026terminalbench,harbor022},
a general evaluation harness that standardizes how agent systems receive tasks
and how their outputs are graded. Harbor runs the agent in an isolated Docker
environment and invokes the grader in a separate, network-disabled container.
When the run ends, only the editable solution and its auxiliary Lean files are
transferred to the verifier; all other agent-side changes are discarded. The
verifier inserts those files into a fresh copy of the original workspace and
runs \textsc{Comparator}~\citep{comparator}. Comparator checks that the submitted
results prove exactly the original statements, do not alter definitions fixed
by the benchmark, use only permitted axioms, and are accepted by Lean.
Successful proofs are additionally replayed by Nanoda~\citep{nanoda}, an
independent proof-checking kernel.
This strict protocol is needed to prevent models from cheating by leveraging their knowledge of Lean syntax, e.g. by using special keywords to trivialize the task by redefining key mathematical objects.
Appendix~\ref{app:evaluation-harness} describes the workspace, budgets, tools,
isolation procedure, and Harbor packaging in detail.

\begin{figure}[H]
  \centering
  \vspace{-5pt}
  \includegraphics[width=\textwidth]{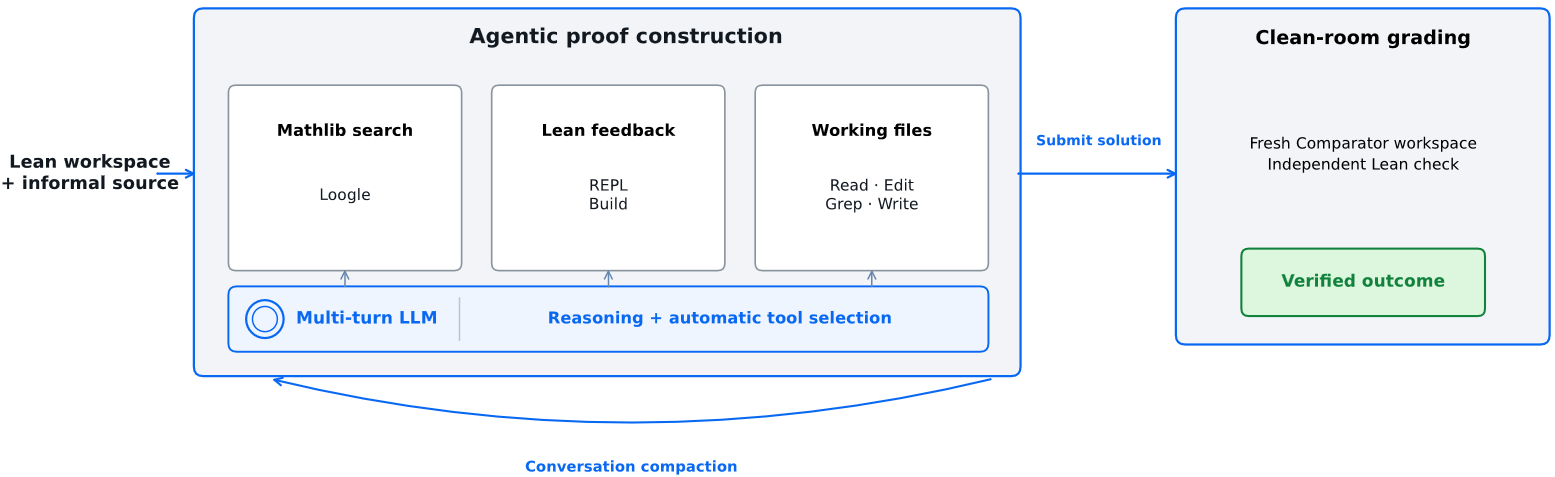}
  \vspace{-15pt}
  \caption{AutoformBench evaluation protocol. A multi-turn model edits an
  isolated Lean workspace using library search, executable Lean feedback, and
  file tools. Submitted solutions are independently checked in a fresh
  Comparator workspace; long runs continue through conversation compaction.}
  \label{fig:evaluation-harness}
\end{figure}

\paragraph{Agentic harness.}
Advanced formalization naturally requires an agentic workflow: the solver must
navigate an unfamiliar development, search for existing library results,
experiment with candidate proof fragments, interpret compiler feedback, and
iteratively extend its solution. We therefore use a lightweight multi-turn
harness. As shown in Figure~\ref{fig:evaluation-harness}, we provide three types
of tools:
\begin{itemize}
  \item \textbf{Filesystem tools} for navigating the workspace and editing
        submission files.
\item \textbf{Lean-execution tools} that allow the model to test partial or complete candidate solutions against the Lean kernel and check their correctness.
  \item \textbf{Mathlib search} through Loogle~\citep{loogle}, for finding
        declarations in the pinned Mathlib library.
\end{itemize}
Tool execution is network-disabled, so the model cannot retrieve external
solutions or documentation during a run.

The workspace serves as durable state throughout the trajectory. When the
conversation approaches the model's context limit, the harness compacts the
interaction history into a summary while preserving the files and running tool
environment. This allows the model to continue working across multiple context
windows and enables test-time scaling through longer trajectories. Each run has a flat provider-cost cap of \$200 per problem. We derive
reported dollar costs from provider-reported token counts using the public API
prices for each model at the time of evaluation. A run terminates when
the model calls \texttt{submit\_report} or exhausts its budget.

\section{Main Results}

We evaluate OpenAI's GPT-6 Astra and GPT-5.6 Sol
\citep{openai2026gpt6astra,openai2026gpt56sol}; Anthropic's Claude Opus~5,
Claude Opus~4.8, Claude Sonnet~5, and Claude Fable~5 (Fable~5 below)
\citep{anthropic2026opus5,anthropic2026opus48,anthropic2026sonnet5,anthropic2026fable5};
Google DeepMind's Gemini~3.7 Flash \citep{googledeepmind2026gemini37flash};
and DeepSeek V4.1 Flash and Kimi K3
\citep{deepseekai2026v41flash,kimiteam2026k3}.
The headline evaluation uses
\texttt{xhigh} reasoning effort for GPT-5.6 Sol;
\texttt{high} for Gemini~3.7 Flash, Claude Sonnet~5, and Claude Opus~4.8; and
\texttt{max} for Claude Opus~5, Fable~5, DeepSeek V4.1 Flash, Kimi K3, and
GPT-6 Astra. A turn is one model response plus its tool-call
batch. Figure~\ref{fig:main-results} summarizes their verified performance with
the \$200 budget cap. Appendix~\ref{app:reasoning-effort} examines sensitivity
to lower effort settings, while Appendix~\ref{app:evaluation-cost} gives the full cost breakdown.
Note that the \$200 per-task budget used for the headline scores is a ceiling rather than a
requirement for evaluating AutoformBench. Evaluators can choose a smaller fixed
budget and still obtain a directly comparable leaderboard. We believe that this is an important point: cost efficiency is essential for the long-term impact of autoformalization, and formalizing the mathematical literature at well as the production of new research at scale will require models and inference harnesses that achieve strong performance at low computational cost. As such, AutoformBench can provide a natural benchmark for measuring progress under explicit cost constraints.

\paragraph{Budget-conditioned leaderboard.}
The tasks evaluated are extremely challenging, and most models require a very long context to solve them. Moreover, models can test their candidate solutions using Lean tools, which allows them to keep iterating until a correct solution is found. As such, the most informative metric is not the absolute success rate of a given model, but rather the number of problems solved given a certain budget (of tokens, turns, etc.).

\begin{figure}[H]
  \centering
  \includegraphics[width=\textwidth]{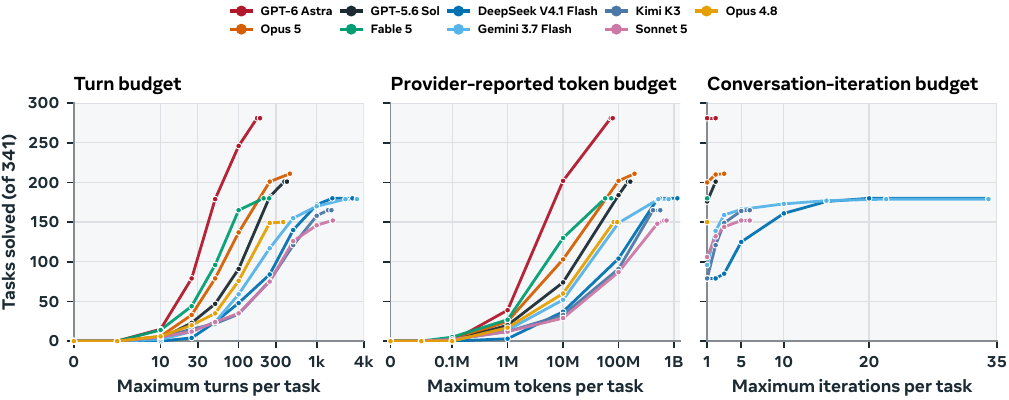}
  \caption{\footnotesize Budget-conditioned leaderboard on the 341 tasks by
  agent turns, provider-reported input plus output tokens, and conversation
  iterations (each context compaction starts a new iteration). Astra is highest
  from 10 turns, approximately 0.41 million total tokens, and the first
  conversation iteration onward. The corresponding dollar-budget view appears
  in Figure~\ref{fig:main-results}.}
  \label{fig:budget-conditioned-leaderboard}
\end{figure}
A verified task enters the curve once its recorded run reaches that resource
level; failed tasks never enter. Figure~\ref{fig:budget-conditioned-leaderboard}
reports this score for three non-dollar per-task resource limits. At any fixed cutoff, the
highest curve identifies the model with the greatest observed verified coverage
under that constraint.
We observe that our set of tasks is particularly challenging, with most frontier models requiring hundred of millions of read and written tokens to reach a 50\% completion rate.
A striking finding is that the efficiency frontier is dominated by two models embodying radically opposed design philosophies: Astra, a superlative frontier model which trades a high cost per token for a very high token-efficiency, and DeepSeek V4.1 Flash, an inference-optimized model with a very low cost per token.

\begin{figure}[H]
  \centering
  \includegraphics[width=0.75\textwidth]{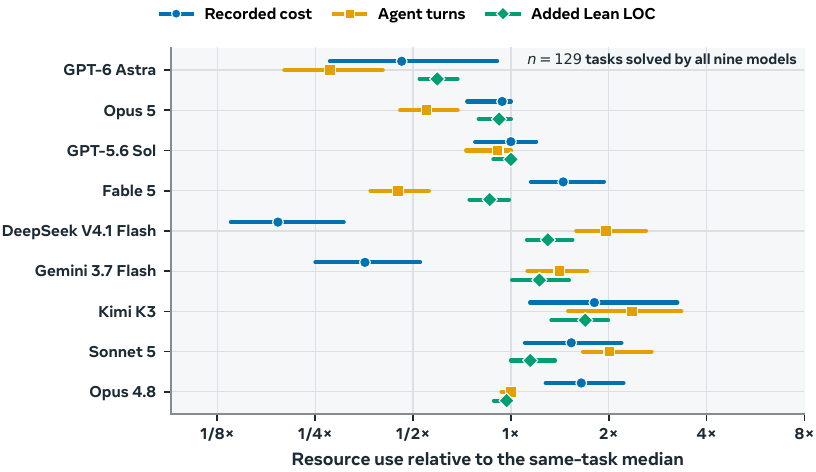}
  \caption{ Compared efficiency on the 129 tasks solved by all nine
models. For each task, cost, turns, and added Lean LOC are
  normalized by the median across models. Points show each model's median ratio
  and horizontal bars show the interquartile range.}
  \label{fig:matched-efficiency-summary}
  \vspace{-10pt}
\end{figure}

\paragraph{Efficiency comparisons.}
Because models solve different task sets, we also compare them on the 129 tasks
solved by all nine models. For each task and resource, we divide
each model's value by the median across models before summarizing the resulting
ratios in Figure~\ref{fig:matched-efficiency-summary}. DeepSeek has the lowest
median recorded cost ($0.19\times$ the same-task median), whereas Astra uses the
fewest turns ($0.28\times$) and produces the most concise answers ($0.59\times$). The
separation between these dimensions shows that monetary cost, interaction
length, and implementation size capture distinct aspects of agent efficiency.
Appendix~\ref{app:matched-efficiency} gives the task-level distributions and
taskwise-best comparison for this shared 129-task cohort.

\paragraph{What makes a problem difficult?}
The nine models evaluated exhibit a strongly stratified difficulty frontier.
Across all nine models, 129 tasks are solved by every model and 59 by none; 153
produce disagreement (Figure~\ref{fig:coverage-histogram}). Across all 36
possible model pairs, the higher-scoring model solves at least 95\% of the
lower-scoring model's successes in 30 pairs. For the median pair, only two tasks
are solved by the lower-scoring model but not by the higher-scoring model. Astra
alone solves 281 tasks and uniquely solves 66 that every other model misses;
conversely, only one task is solved by another model and missed by Astra. This
strong nesting indicates a largely shared difficulty frontier.

\begin{figure}[h]
  \centering
  \includegraphics[width=0.65\textwidth]{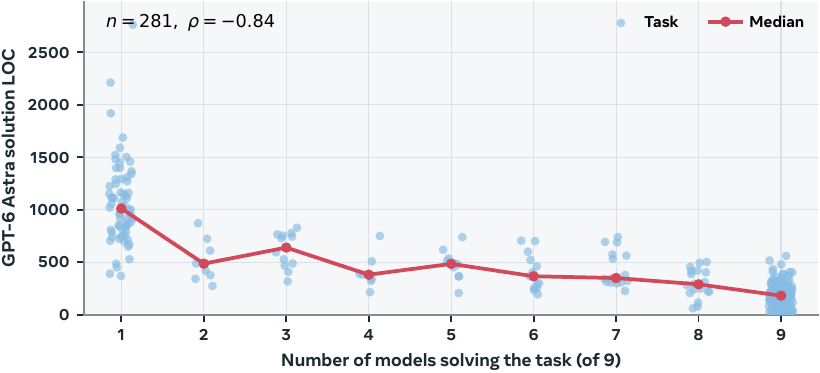}
  \caption{GPT-6 Astra solution LOC versus the number of models solving each
  task, restricted to Astra's 281 verified successes. Each blue point is one
  task; the red line connects the median
  solution LOC at each solver count.}
  \label{fig:astra-loc-vs-coverage}
\end{figure}

Two related measures track this difficulty. Across the nine models, mean success
falls from 66.4\% on single-hole tasks to 32.5\% on multi-hole tasks.
Figure~\ref{fig:astra-loc-vs-coverage} compares GPT-6 Astra's solution LOC with
the number of models that solve the same task. Across Astra's 281 verified
solutions, the correlation is $\rho=-0.84$: longer solutions tend to correspond
to tasks solved by fewer models. By contrast, we find little evidence of
specialization by mathematical field; Appendix~\ref{app:book-results} reports
the per-textbook results.

\section{Reliability and Difficulty Analysis}

\paragraph{Agent-Built Infrastructure.}

Solving AutoformBench typically requires building a local formal library around
the target theorem. Among the 281 accepted GPT-6 Astra solutions, 92.9\% use
auxiliary lemmas, 73.3\% introduce new definitions, and 94.0\% of
dependency-reachable proof code lies in supporting declarations rather than the
graded proof bodies.
This remains true for tasks exposing only one proof hole: 90.4\% use auxiliary lemmas, and supporting declarations account for 93.4\% of live proof code. A single-hole task is therefore not necessarily an isolated theorem-proving problem.
Appendix~\ref{app:agent-infrastructure} gives the complete methodology and
distributions.

\paragraph{Run-to-Run Stability.}
\label{sec:stability-runs}

\begin{figure}[h]
  \centering
  \includegraphics[width=0.96\textwidth]{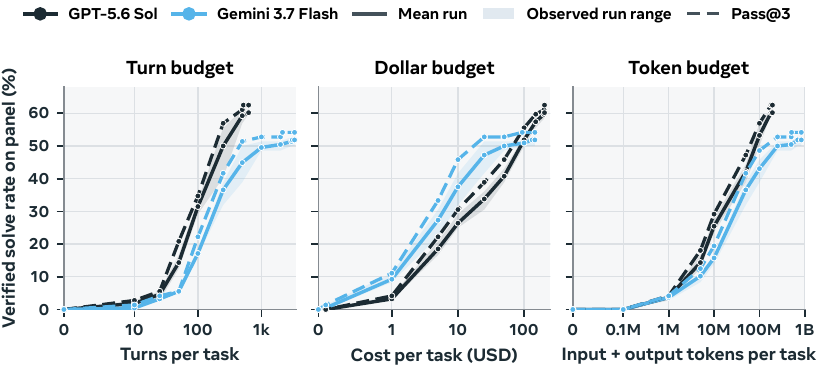}
  \caption{Verified coverage on the 72-problem panel as turn, recorded-cost,
  and provider-reported input-plus-output-token budgets increase.}
  \label{fig:stability-budget}
\end{figure}

Because AutoformBench permits long and expensive agent trajectories, the
full-benchmark leaderboard reports one scored trajectory per model and problem
(pass@1). To test whether these results are sensitive to run-to-run
variation, we perform a smaller repeatability study: GPT-5.6 Sol and
Gemini~3.7 Flash are each run three times on a fixed, stratified panel of 72
problems spanning all 24 textbooks. Appendix~\ref{app:stability-panel} gives the
complete panel construction and repeated-run protocol.
Figure~\ref{fig:stability-budget} shows both quantities as the available turn,
cost, and token budgets increase.

Outcomes are highly stable. For each model, 69/72 problems receive the same
result in all three runs. Sol solves 43--44 problems per run and 45 under
pass@3; Gemini solves 36--38 per run and 39 under pass@3. Thus pass@3 adds only
one solution beyond the best individual run for either model.

\section{Ablations}
\label{sec:ablations}

\paragraph{Search and Executable Feedback}
To measure the effect of tool access, we first remove the agent loop entirely.
On the 230 workspaces containing exactly one proof hole, GPT-5.6 Sol and
Gemini~3.7 Flash receive the complete problem-local context and informal source
in a single request, but cannot use tools, receive compiler feedback, or revise
their answer. We use the headline model settings and the same clean-room grader.
They solve only 6/230 (2.6\%) and 3/230 (1.3\%) problems, respectively, compared
with 163/230 (70.9\%) and 148/230 (64.3\%) under the full agentic evaluation.

We then remove individual tools while retaining the agent loop. To control
cost, these experiments reuse the fixed 72-problem panel from
Section~\ref{sec:stability-runs}. We remove the REPL, Loogle, and full-project
build individually and in combination; every experimental setup retains filesystem
tools, report submission, the informal source, and the \$200 per-problem budget.
Table~\ref{tab:tool-ablation} reports the results.

\begin{table}[H]
  \centering
  \scriptsize
  \setlength{\tabcolsep}{3.5pt}
  \begin{tabular}{lrr}
    \toprule
    Condition & Gemini~3.7 Flash & GPT-5.6 Sol \\
    \midrule
    Full tools, three runs & 38, 36, 38 & 43, 44, 43 \\
    No REPL                & 36         & 45 \\
    No Loogle              & 35         & 44 \\
    No build               & 24         & 39 \\
    No REPL or build       & 1          & 4 \\
    Files only             & 1          & 1 \\
    \bottomrule
  \end{tabular}
  \caption{Verified successes under tool ablations on the fixed 72-problem
  panel. The full-tool row reports all three runs; other conditions use one run.}
  \label{tab:tool-ablation}
\end{table}

Removing either the REPL or Loogle alone leaves success rates close to those
observed across the three full-tool runs. Removing build has a larger effect,
particularly for Gemini. Most strikingly, removing both executable Lean-feedback
interfaces collapses performance from 36--45 solved problems to only 1 for
Gemini and 4 for GPT-5.6 Sol. With only file operations remaining, both models
solve just 1/72 problems. Thus structured Mathlib search through Loogle is not
essential when filesystem search remains available, and the REPL and build seem to be to some extent substitutable. Appendix~\ref{app:ablation-details} gives the detailed tool-use and
fallback analysis.

\paragraph{Context-Iteration Scaling}

We measure how much the compaction mechanism described in
Section~\ref{sec:evaluation-protocol} contributes to task completion by
comparing coverage at the end of the initial context window with final coverage
after subsequent context windows. DeepSeek solves 101 additional tasks after
its first context window, Kimi 86, Gemini 83, and Sonnet 46. The gains are
smaller for Opus and GPT-5.6 Sol, while every successful Astra, Fable, and
Opus~4.8 trajectory finishes within its initial context window.

\begin{table}[H]
  \centering
  \scriptsize
  \setlength{\tabcolsep}{3.5pt}
  \begin{tabular}{lrrr}
    \toprule
    Model & Before compaction & Final & Gain \\
    \midrule
    GPT-6 Astra          & 281 & 281 & 0 \\
    Opus~5              & 200 & 211 & 11 \\
    Fable~5             & 181 & 181 & 0 \\
    GPT-5.6 Sol         & 176 & 201 & 25 \\
    Opus~4.8            & 150 & 150 & 0 \\
    Sonnet~5            & 106 & 152 & 46 \\
    Gemini~3.7 Flash    & 96  & 179 & 83 \\
    DeepSeek V4.1 Flash & 79  & 180 & 101 \\
    Kimi K3             & 79  & 165 & 86 \\
    \bottomrule
  \end{tabular}
  \caption{Verified coverage before the first conversation compaction and after
  all recorded context iterations.}
  \label{tab:compaction-scaling}
\end{table}
\paragraph{Agent-Scaffold Comparison}

We evaluate Opus~5 through both our common agent loop and Claude
Code~\citep{claudecode2026}. Both configurations receive the same 341 formal
workspaces and textbook context, have a maximum budget of \$200 per task, and
undergo identical clean-room verification. Claude Code uses the benchmark's
Lean REPL, Loogle, build, and report services alongside its native
\texttt{Read}, \texttt{Edit}, \texttt{Write}, \texttt{Glob}, and \texttt{Grep}
tools; Bash, web access, and subagents are disabled.

\begin{figure}[H]
  \centering
  \includegraphics[width=0.73\textwidth]{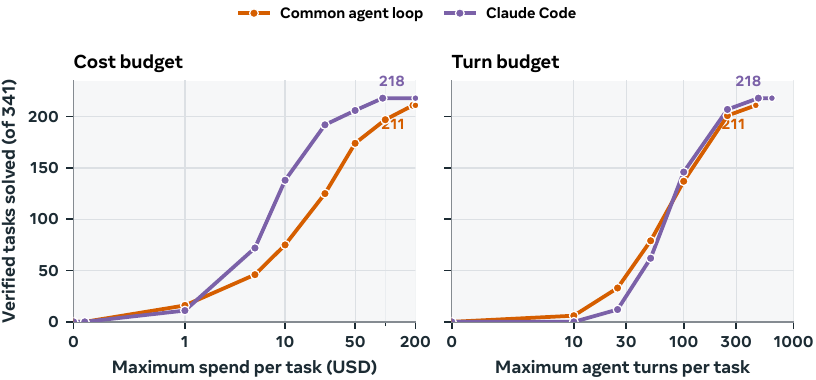}
  \caption{Budget-conditioned comparison of Opus~5 under the common agent loop
  and Claude Code on the same 341 tasks.}
  \label{fig:claude-code-opus-budget}
\end{figure}

Claude Code solves 218/341 tasks, compared with 211/341 for the common loop.
The configurations agree on 308 problems: both solve 198, Claude Code alone
solves 20, the common loop alone solves 13, and neither solves 110.
Figure~\ref{fig:claude-code-opus-budget} compares their coverage over recorded
dollar and turn budgets.

Claude Code's advantage is primarily one of efficiency and persistence rather
than a different mathematical strategy. In 18 of its 20 exclusive successes,
both configurations pursue the same mathematical approach, but the common loop
stops before completing the final obligations. Plausible explanations include its native system prompt, internal
compaction policy, and training for its own tools, none of which we can observe
or isolate.
Appendix~\ref{app:ablation-details} provides the detailed tool-use and failure analysis.

\paragraph{Informal-Source Ablation}

\begin{figure}[H]
  \centering
  \includegraphics[width=0.95\textwidth]{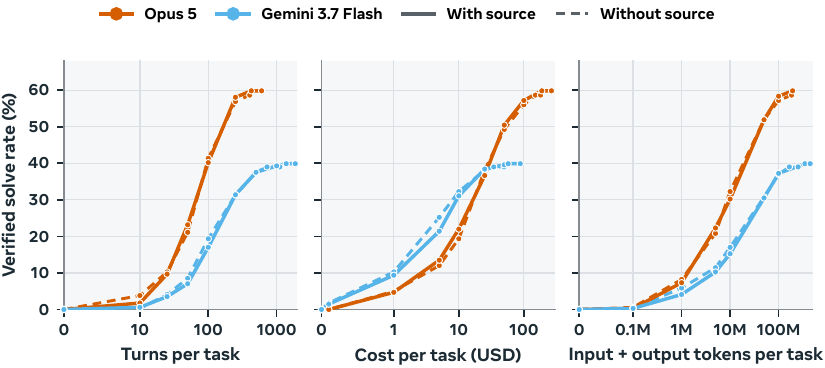}
  \caption{Efficiency by turns, recorded provider cost, and provider-reported
  input plus output tokens with and without the textbook source on the 341
  audited benchmark tasks for Opus~5 and Gemini~3.7 Flash.}
  \label{fig:source-ablation-efficiency}
\end{figure}

We ablate access to the textbook by evaluating Opus~5 and Gemini~3.7 Flash on
all 341 tasks with the same formal workspace, model settings, tools, budget,
stopping rule, and grader. Surprisingly, although both models consult the
textbook in more than 93\% of source-enabled runs, it provides little aggregate
improvement: Opus solves 204 tasks with the source and 200 without it, while
Gemini solves 136 and 133, respectively. Neither paired difference is
statistically significant, and turns, cost, and token use are similar across
conditions (Figure~\ref{fig:source-ablation-efficiency}). The source also does
not systematically shorten the generated Lean code on tasks solved in both
conditions. Trace inspection instead shows a task-specific effect: some runs
benefit from the book's proof decomposition, whereas some source-free runs find
shorter arguments better adapted to Lean. Thus explicit source access changes
individual proof trajectories but provides no clear aggregate advantage once
the formal statement and its local context are available. We attribute this to the strong mathematical knowledge of current frontier models.
Appendix~\ref{app:source-cases} gives the complete paired statistics,
source-use analysis, and case studies.

\section{Limitations and Conclusion}

\paragraph{Limitations.}
AutoformBench covers 24 mathematical sources, but it is not representative of
all advanced mathematics: its domains and statements are constrained by what
can presently be expressed against Mathlib within a bounded evaluation. The
benchmark supplies expert-written formal statements, so it evaluates proof
completion and construction of supporting theory rather than end-to-end
statement autoformalization.
Finally, kernel acceptance establishes a theorem
relative to its audited statement and assumptions; it does not establish that
the generated code has the abstraction quality or maintainability required for
upstream inclusion in a useful library  such as Mathlib.

\paragraph{Conclusion.}
AutoformBench is a natural step toward measuring how well models can construct
missing formal infrastructure. Its problems, however, remain isolated in
separate workspaces. The next step is to test whether agents can recognize
shared missing theory, build reusable abstractions, and amortize that work
across related results. Moving from isolated problems to cumulative developments
would measure the transition from theorem solving to extending formal libraries.

\clearpage

\section*{Acknowledgments}
JK and NP thank the Simons Foundation for support through the Collaborative Grant “The Physics of Learning and Neural Computation” as well as support by the NSF through NRT Award 1922658.

AH benefited from Hi! PARIS and State funding managed by the French National
Research Agency (ANR) under the France 2030 program, reference
ANR-23-IACL-0005.
The authors declare no competing interests.

\section*{Reproducibility Statement}
We release all 341 tasks in Harbor format, each pinned to Lean 4.32.2 and a fixed Mathlib commit and including the graded-declaration and permitted-axiom specification and informal source material. Appendix B details the tools, sandbox, complete prompts, budget and termination rules, clean-room grading procedure, and exact revisions of Comparator, Nanoda, Landrun, and Loogle; model settings appear in Section 5, and the repeatability panel and its selection seed in Appendix H. Since proprietary API models may change, trajectories may not be exactly reproducible, but any submission can be re-verified independently from its Lean files alone.

\clearpage

\bibliography{references}

\clearpage
\beginappendix

\section{Benchmark Structure: Proof Obligations and Local Context}
\label{app:benchmark-structure}
\label{app:hole-distribution}

This section gives two complementary views of the benchmark : the number of explicit proof obligations in each workspace and
the local formal context on which the target statements depend.

\subsection{Proof Obligations}

Table~\ref{tab:hole-distribution} gives the complete distribution of graded
proof obligations across the frozen benchmark. Of the 1{,}155 graded
declarations, 1{,}147 use theorem syntax and 8 use instance syntax for
proposition-valued typeclasses.

\begin{table}[H]
  \centering
  \scriptsize
  \setlength{\tabcolsep}{3.5pt}
  \begin{tabular}{rrrrrrrr}
    \toprule
    Holes & Tasks & Share & Oblig. & Holes & Tasks & Share & Oblig. \\
    \midrule
     1 & 230 & 67.45\% & 230 & 14 & 4 & 1.17\% & 56 \\
     2 &  25 &  7.33\% &  50 & 15 & 1 & 0.29\% & 15 \\
     3 &  19 &  5.57\% &  57 & 16 & 1 & 0.29\% & 16 \\
     4 &   9 &  2.64\% &  36 & 17 & 1 & 0.29\% & 17 \\
     5 &   7 &  2.05\% &  35 & 18 & 2 & 0.59\% & 36 \\
     6 &   5 &  1.47\% &  30 & 19 & 1 & 0.29\% & 19 \\
     7 &   2 &  0.59\% &  14 & 20 & 2 & 0.59\% & 40 \\
     8 &   4 &  1.17\% &  32 & 24 & 1 & 0.29\% & 24 \\
     9 &   3 &  0.88\% &  27 & 26 & 2 & 0.59\% & 52 \\
    10 &   5 &  1.47\% &  50 & 27 & 2 & 0.59\% & 54 \\
    11 &   4 &  1.17\% &  44 & 30 & 1 & 0.29\% & 30 \\
    12 &   5 &  1.47\% &  60 & 33 & 1 & 0.29\% & 33 \\
    13 &   3 &  0.88\% &  39 & 59 & 1 & 0.29\% & 59 \\
    \midrule
    \multicolumn{8}{c}{341 tasks; 1{,}155 graded proof obligations} \\
    \bottomrule
  \end{tabular}
  \caption{Number of graded proof holes per benchmark task. \textbf{Oblig.}
  is the total number of obligations contributed by that row.}
  \label{tab:hole-distribution}
\end{table}

The 111 multi-hole tasks have a median of 5 and a mean of 8.33 holes. The
distribution is concentrated: the largest 10 tasks contain 292/1{,}155
obligations (25.3\%), the largest 20 contain 451 (39.0\%), and the 44 tasks with
at least eight holes contain 703 (60.9\%). Designating the final graded
declaration in each workspace as its headline target leaves 814 supporting
obligations, or 70.5\% of all graded declarations. Agent-added definitions and
lemmas are additional to these benchmark-provided holes.

\subsection{Formal Context Required to State a Task}

A formal target uses constants from both Mathlib and its benchmark workspace.
To quantify the context supplied before proof search, we start from each
elaborated target type and recursively follow references into
\texttt{Challenge.lean} and \texttt{ChallengeDeps.lean}; for a multi-target task,
we take the union of the resulting graphs. Table~\ref{tab:statement-infrastructure}
reports the number of reachable declarations, their source lines of code, and
the longest local dependency chain. No model-generated code is counted. We use
the 281 tasks solved by GPT-6 Astra so that these input-side measurements can be
paired with the accepted-proof analysis in
Appendix~\ref{app:agent-infrastructure}.

\begin{table}[H]
  \centering
  \small
  \begin{tabular}{lrrr}
    \toprule
    Input-side metric & Median & IQR & Maximum \\
    \midrule
    Reachable workspace declarations & 5 & 2--12 & 54 \\
    Source LOC of those declarations & 22 & 7--49 & 293 \\
    Longest workspace dependency chain & 2 & 1--3 & 9 \\
    \bottomrule
  \end{tabular}
  \caption{Formal context required by the graded target types in the 281-task
  GPT-6 Astra accepted-solution cohort.}
  \label{tab:statement-infrastructure}
\end{table}

The median task depends on five workspace declarations occupying 22 lines, with
a longest local chain of two. Among constants mentioned directly in its target
types, the median Mathlib fraction is 65.0\%. Thus predominantly Mathlib
vocabulary can coexist with a nontrivial local API. These measurements describe
context supplied with the task, not infrastructure built by the agent.

\section{Execution and Clean-Room Verification}
\label{app:evaluation-harness}

\subsection{Workspace Structure}

\textsc{Lean Eval}~\citep{lean_eval} generates a standalone Comparator workspace
for each benchmark item.
All workspaces use Lean~4.32.2 and Mathlib commit
\texttt{905b95818eb32af7874a58b427f50c1711a5e96c}.
The principal files are:

\begin{verbatim}
ChallengeDeps.lean       supporting definitions
Challenge.lean           trusted target declarations
Submission.lean          agent's candidate proofs
Submission/Helpers.lean  optional auxiliary lemmas
Solution.lean            trusted grading/build wiring
config.json              graded names and permitted axioms
book/                    informal textbook material
TASK.md                  agent instructions
\end{verbatim}

The agent receives a disposable copy of this workspace. Although it can inspect
the trusted challenge and build configuration, its only graded deliverables are
\texttt{Submission.lean} and the optional \texttt{Submission/} directory.
\texttt{Submission.lean} is initialized with the target declarations and
\texttt{sorry} proof bodies; agents may add definitions and auxiliary lemmas but
must fill every target proof.

\subsection{Tools and Execution Sandbox}

The agent receives five classes of MCP service: scoped filesystem operations;
\texttt{run\_lean\_code}, a scratch Lean REPL using the benchmark's pinned
Mathlib environment; \texttt{loogle\_search}, for searching Mathlib declarations;
a fixed build operation equivalent to \texttt{lake build Solution}; and
\texttt{submit\_report}. The build service exposes compiler diagnostics without
granting a general shell. The report service records a structured outcome
(\texttt{solved} or \texttt{unsolved}), optional diagnostic flags, and a short
free-text account of progress or failure. It is a source of diagnostic metadata,
not the grading oracle.

Each MCP subprocess runs under Bubblewrap. The problem workspace is its only writable
mount; required system, Lean, and tool directories are mounted read-only; temporary
storage and the home directory are private; the environment is allowlisted; and
network access is disabled.

\subsection{Agent Prompts}

API-model runs use a fixed solver prompt. The complete machine-readable
templates are released with the evaluation artifacts; we
reproduce the result-relevant instructions below, omitting the diagnostic-report
flag taxonomy.

\paragraph{Solver instruction.}
\begin{quote}\small
You are an expert Lean~4 theorem-proving agent working in a self-contained
workspace. Read \texttt{TASK.md} and \texttt{Challenge.lean}, then replace every
\texttt{sorry} in \texttt{Submission.lean}; auxiliary definitions and lemmas may
be added to \texttt{Submission/Helpers.lean}. Use the filesystem tools to edit the
submission, \texttt{run\_lean\_code} as a scratchpad, \texttt{loogle\_search} to
search Mathlib, and \texttt{build} to compile the candidate. Missing Mathlib
infrastructure is part of the task: build the necessary supporting theory rather
than stopping when a lemma cannot be found. Do not modify target statements,
introduce axioms, or bypass proof checking. Finish by calling
\texttt{submit\_report} with a solved or unsolved outcome and a concise account of
the work completed and any remaining obstacle.
\end{quote}

\paragraph{Continuation prompt.}
If an agent stops without a report, it is reminded that it must either submit a
completed proof or continue using the available tools. If it reports
\texttt{unsolved} before reaching the minimum-effort threshold (10\% of its budget), it receives:
\begin{quote}\small
You have not worked hard enough on this problem yet. These problems are
intentionally difficult and may require substantial new Lean infrastructure;
missing Mathlib lemmas are expected, not a reason to stop. Spend more time
developing the necessary definitions and helper lemmas, test them with the
available tools, and keep iterating until the solution builds.
\end{quote}

\paragraph{Conversation compaction.}
To support trajectories longer than one context window, the agentic loop
compacts its conversation after a completed model turn when the context
approaches its limit.

For compaction, the harness calls the same model at the same reasoning setting, with tools disabled, using the following prompt:

\begin{quote}\small
\textbf{System instruction.} You are a summarizer. Produce a concise summary of
the conversation below. Focus on: key findings, decisions made, actions taken,
and remaining work. Be specific about important details like file paths, names,
and values.

\medskip
\textbf{User template.}\par
\texttt{Summarize this conversation:}\par
\smallskip
\texttt{\{conversation\_text\}}
\end{quote}

The original solver system prompt is then restored, and the prior non-system
history is replaced by one user message prefixed
\texttt{[Context summary of previous conversation]}. Compaction changes neither the workspace files nor the running tool services; thus
verified Lean progress survives even though the detailed dialogue does not. If
summarization fails or returns no usable text, the original history is restored. The
procedure may repeat within a run.

\subsection{Budget and Termination}
All models are evaluated under a maximum provider-cost budget of
\$200 per problem. A solved report may terminate immediately.
An unsolved report submitted before 10\% of the budget is normally rejected and
the agent is instructed to continue; the local-model driver accepts the sixth
such report to prevent an unproductive rejection loop. Appendix~\ref{app:persistence}
analyzes the effect of this rule. The final 10\% is reserved for at most three wrap-up turns so that an agent
that exhausts its proving budget can still record partial progress and remaining
obstacles. Reported dollar costs are computed from provider-reported input,
output, and cached-token counts, including separate cache-read and cache-write
counts where available, using the public API prices applicable to each model at
evaluation time. They measure estimated API-equivalent inference cost rather
than hardware-normalized compute or invoiced expenditure.

\subsection{Clean-Room Grading}
Many frontier models have a tendency to cheat, which makes rigorous grading a crucial necessity.
When an agent finishes, the harness creates a new workspace from the trusted
benchmark copy and overwrites exactly two paths with agent output:
\texttt{Submission.lean} and \texttt{Submission/}. All other files---including
the reference declarations, build wiring, graded-name list, pinned definitions,
and permitted-axiom configuration---come from the trusted copy. Consequently,
changes made elsewhere in the disposable workspace cannot affect the verdict.

Comparator then compiles the candidate in its Landrun sandbox~\citep{landrun} and checks that
each selected declaration has the same statement as its trusted counterpart,
that pinned definitions retain their values, and that the proof uses only
permitted axioms. Accepted submissions are additionally replayed through the
independent Nanoda kernel~\citep{nanoda}. The resulting Comparator verdict, rather than the
agent's \texttt{submit\_report} claim or its local build result, defines
benchmark success.

\paragraph{Harbor task format and release.}
Harbor's container-based task format and execution harness, also used by
Terminal-Bench~2.0, provide a common interface for heterogeneous agent
scaffolds. The public release packages every benchmark workspace as a
Harbor~0.22.0 task, allowing any Harbor-supported or custom agent to be evaluated
under the grading semantics described above.

\section{Evaluation Cost}
\label{app:evaluation-cost}

\begin{figure}[H]
  \centering
  \includegraphics[width=\textwidth]{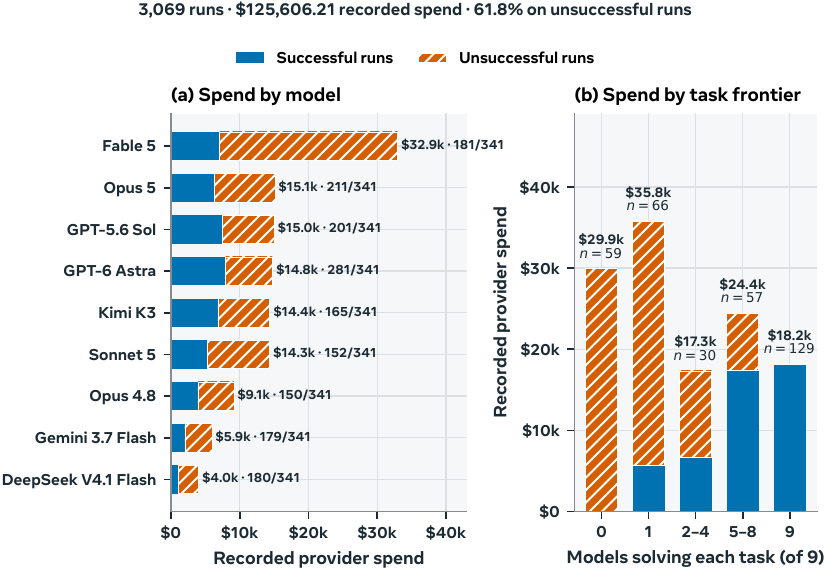}
  \caption{Recorded provider spending across the nine leaderboard
  campaigns. Left: spend by model, split between verified and unsuccessful
  runs; labels give total spend and verified score. Right: the same spending
  grouped by the number of models that solve each task; $n$ is the number of
  tasks in each group.}
  \label{fig:evaluation-cost}
\end{figure}

\paragraph{Evaluation spending and the unsolved frontier.}
Figure~\ref{fig:evaluation-cost} separates recorded provider spending by model,
run outcome, and cross-model task coverage. Across all 3{,}069 model--task runs,
the 1{,}369 unsuccessful runs account for \$77{,}676.70 of the \$125{,}606.21
total (61.8\%). The 59 tasks solved by none of the nine models consume
\$29{,}920.86 and 178{,}744 agent turns. The 66 tasks solved by exactly one model
consume a further \$35{,}757.19; together, these 125 frontier tasks account for
\$65{,}678.05, or 52.3\% of all recorded leaderboard spending. Thus more than half
of the evaluation cost is devoted to tasks that remain unsolved or lie at the
edge of current model capability.

\section{Matched Model Efficiency}
\label{app:matched-efficiency}

\paragraph{Controlled comparison on common successes.}
To separate model behavior from differences in the tasks each model solves, we
restrict to the 129 tasks solved by all nine models. For every
task and resource, we divide each model's recorded value by the minimum among
the nine models, then report the median ratio in
Table~\ref{tab:matched-efficiency}. A ratio of one is best. The final three
columns count tasks on which a model attains the minimum, counting ties for each
model. DeepSeek is cheapest on 85/129 tasks. Astra uses the fewest turns on
109/129 and leaves the least added code on 107/129.

\begin{table}[H]
  \centering
  \scriptsize
  \setlength{\tabcolsep}{3.5pt}
  \begin{tabular}{lrrrrrr}
    \toprule
    & \multicolumn{3}{c}{Median ratio to taskwise best} &
      \multicolumn{3}{c}{Tasks attaining best} \\
    \cmidrule(lr){2-4}\cmidrule(lr){5-7}
    Model & Cost & Turns & Added LOC & Cost & Turns & Added LOC \\
    \midrule
    GPT-6 Astra         & 2.70 & 1.00 & 1.00 & 18 & 109 & 107 \\
    Opus~5              & 4.84 & 1.82 & 1.50 & 0  & 2  & 5 \\
    GPT-5.6 Sol         & 5.71 & 3.07 & 1.65 & 1  & 1  & 2 \\
    Fable~5             & 8.57 & 1.52 & 1.47 & 0  & 21 & 13 \\
    DeepSeek V4.1 Flash & 1.00 & 6.92 & 2.31 & 85 & 0  & 0 \\
    Gemini~3.7 Flash    & 2.08 & 4.88 & 2.10 & 23 & 0  & 2 \\
    Kimi K3             & 11.38 & 9.83 & 2.76 & 0 & 1 & 1 \\
    Sonnet~5            & 9.79 & 8.21 & 2.12 & 2  & 2  & 1 \\
    Opus~4.8            & 9.71 & 3.40 & 1.63 & 0  & 2  & 6 \\
    \bottomrule
  \end{tabular}
  \caption{Efficiency fingerprints on the identical 129-task cohort solved by
  all nine models. Added LOC is the net code remaining in the
  final submission. Best-task counts include ties.}
  \label{tab:matched-efficiency}
\end{table}

\begin{figure}[H]
  \centering
  \includegraphics[width=0.94\textwidth]{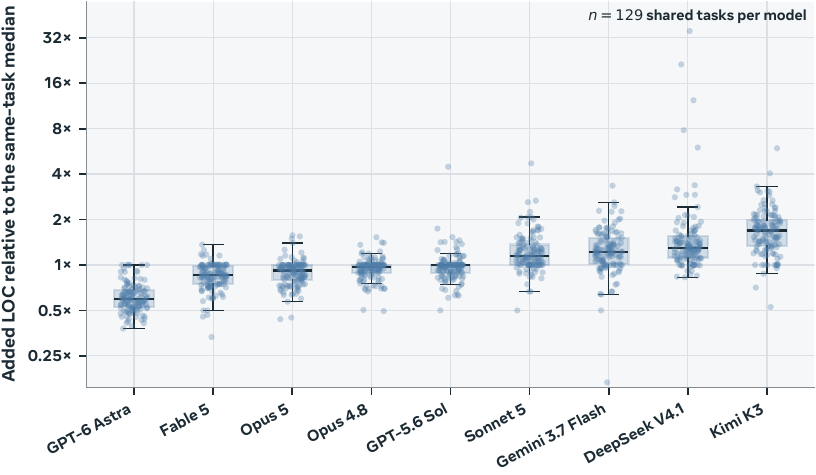}
  \caption{Added Lean code on the 129 tasks solved by all nine
  models, normalized by the median LOC for the same task. Each point is one
  model--task pair; boxes summarize the per-model distributions. A value of
  $1\times$ is the within-task median, while $0.5\times$ and $2\times$ denote
  half and twice that amount of code. Models are ordered by their median ratio.}
  \label{fig:matched-loc}
\end{figure}

\begin{figure}[H]
  \centering
  \includegraphics[width=0.94\textwidth]{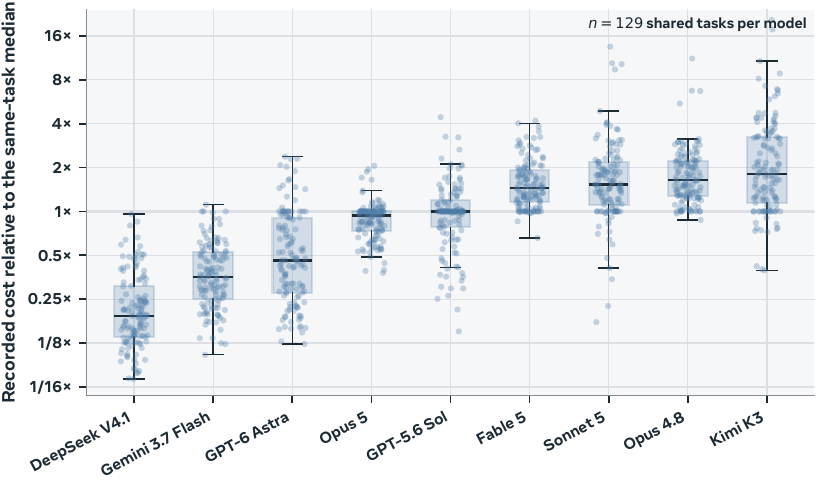}
  \caption{Recorded provider cost on the 129 shared tasks, normalized by the
  median cost for the same task. Each point is one model--task pair; boxes
  summarize the per-model distributions. Models are ordered by median ratio.}
  \label{fig:matched-cost}
\end{figure}

\begin{figure}[H]
  \centering
  \includegraphics[width=0.94\textwidth]{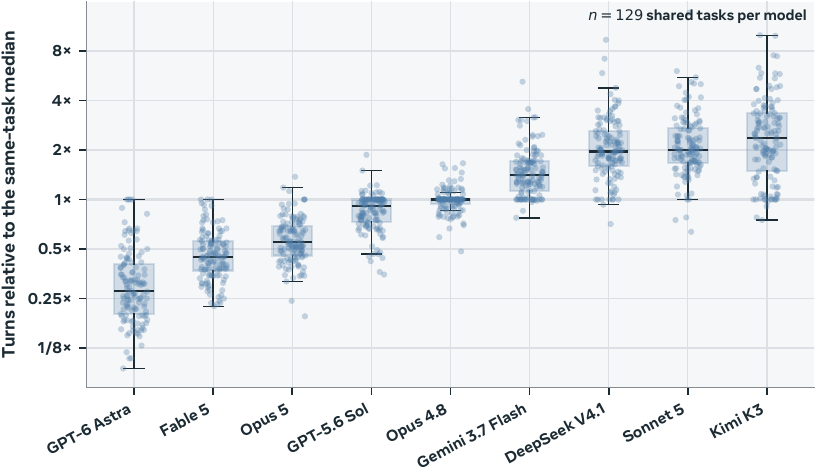}
  \caption{Agent turns on the 129 shared tasks, normalized by the median turn
  count for the same task. Each point is one model--task pair; boxes summarize
  the per-model distributions. Models are ordered by median ratio.}
  \label{fig:matched-turns}
\end{figure}

\paragraph{Matched resource profiles.}
Figure~\ref{fig:matched-efficiency-summary} summarizes the matched comparison in
the main text. Figures~\ref{fig:matched-loc}--\ref{fig:matched-turns} show the
underlying task-level distributions on the same 129 tasks. For each task and
metric, we divide each model's value by the median across the nine models. This
normalization removes the large between-task variation and exposes systematic
differences among models.

\section{Shared Difficulty Frontier}
\label{app:shared-frontier}
\label{app:book-results}

Table~\ref{tab:book-results} reports the complete per-textbook results. Absolute
difficulty varies substantially across sources, but the model outcomes do not
reveal a broad domain-specific specialization that would explain the shared
frontier.

\input{tables/book_results}

\begin{figure}[H]
  \centering
  \includegraphics[width=0.88\textwidth]{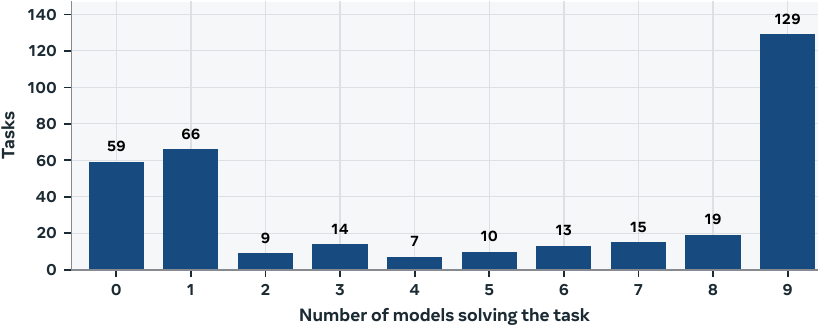}
  \caption{Number of benchmark tasks solved by exactly $k$ of the nine models.}
  \label{fig:coverage-histogram}
\end{figure}

\paragraph{Cross-model coverage.}
Figure~\ref{fig:coverage-histogram} gives the complete distribution of task
coverage across the nine models.

The distribution is strongly polarized: 129 tasks are solved by all nine models,
59 by none, and 153 produce disagreement. Across all 36 model pairs ordered by
total score, the higher-scoring model solves at least 95\% of the lower-scoring
model's successes in 30 pairs; for the median pair, it misses only two such
tasks. Astra solves 281 tasks, while the union of all nine systems reaches only
282.

\section{Agent-Built Infrastructure and Proximity to Mathlib}
\label{app:agent-infrastructure}
\label{app:mathlib-proximity}

\subsection{From Generated Submission to Live Proof}

\begin{figure}[H]
  \centering
  \includegraphics[width=0.68\textwidth]{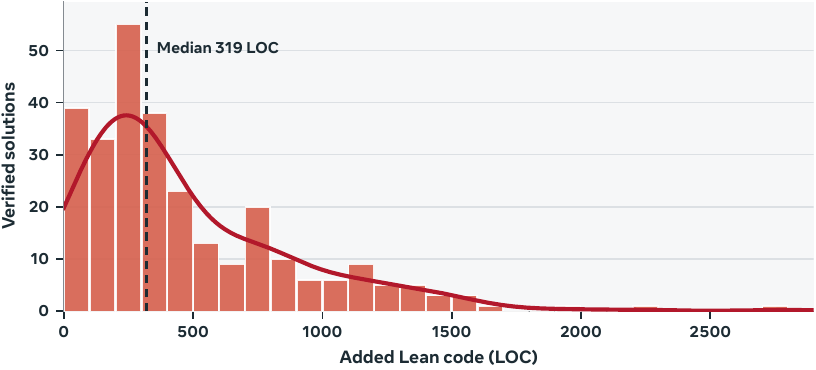}
  \caption{Distribution of skeleton-subtracted, nonblank, noncomment Lean code
  added by GPT-6 Astra across its 281 accepted solutions.}
  \label{fig:solution-loc}
\end{figure}

\paragraph{Accepted-solution cohort and generated code.}
We analyze all 281 clean-room-accepted GPT-6 Astra solutions, comprising 208
single-hole and 73 multi-hole tasks. This is the broadest accepted cohort from
a single model. Before filtering unused code, these solutions
contain a median of 319 skeleton-subtracted, nonblank, noncomment lines of added
Lean; the largest contains 2{,}763. Figure~\ref{fig:solution-loc} shows the full
distribution.

\paragraph{Live dependency closure.}
Raw generated code can include abandoned proof attempts. To isolate the code
that supports an accepted result, we start from every graded proof and
recursively follow dependencies into declarations added by the agent. For a
multi-hole task, we take the union of these closures. Everything unreachable
from a graded proof is excluded from the live measurements.

Table~\ref{tab:live-proof-infrastructure} reports the resulting proof
developments. Auxiliary lemmas occur in 261/281 solutions (92.9\%), new
definitions in 206/281 (73.3\%), and new instances in 68/281 (24.2\%); only
20/281 (7.1\%) use no auxiliary declarations. Supporting declarations
contribute 118{,}853 of 126{,}388 aggregate live LOC (94.0\%), while the graded
target proof bodies contribute 7{,}535 (6.0\%). The live closures retain 95.4\%
of all generated code in aggregate, with a median per-task share of 96.4\%.

\begin{table}[H]
  \centering
  \small
  \begin{tabular}{lrrr}
    \toprule
    Live proof metric & Median & IQR & Maximum \\
    \midrule
    Total live proof LOC & 300 & 182--669 & 2{,}640 \\
    Used helper declarations & 24 & 10--56 & 223 \\
    Used helper LOC & 283 & 151--638 & 2{,}639 \\
    Used definitions & 3 & 0--10 & 77 \\
    Used lemmas & 20 & 9--44 & 179 \\
    \bottomrule
  \end{tabular}
  \caption{Dependency-closed infrastructure in accepted GPT-6 Astra
  solutions. Unused generated declarations are excluded.}
  \label{tab:live-proof-infrastructure}
\end{table}

\subsection{Single- and Multi-Hole Workspaces}

Table~\ref{tab:single-multi-infrastructure} shows that substantial supporting
theory is not confined to workspaces exposing many holes.

\begin{table}[H]
  \centering
  \small
  \begin{tabular}{lrr}
    \toprule
    Metric & Single-hole ($n=208$) & Multi-hole ($n=73$) \\
    \midrule
    Median live proof LOC & 271.5 & 636 \\
    Median used helper declarations & 20 & 53 \\
    Median used definitions & 2 & 9 \\
    Median used lemmas & 16 & 47 \\
    Uses auxiliary lemmas & 90.4\% & 100\% \\
    Uses new definitions & 67.8\% & 89.0\% \\
    Supporting share of aggregate live LOC & 93.4\% & 95.1\% \\
    \bottomrule
  \end{tabular}
  \caption{Dependency-closed proof infrastructure by workspace hole count.}
  \label{tab:single-multi-infrastructure}
\end{table}

Even a single exposed hole therefore usually requires a substantial local
theory: 90.4\% of single-hole successes use auxiliary lemmas, 67.8\% introduce
definitions, and 93.4\% of their live LOC lies outside the final target body.
Conversely, a multi-hole workspace may expose many routine obligations
explicitly, so hole count alone is not a complete measure of difficulty.

\subsection{Resource Use Across Successful Runs}

\begin{figure}[H]
  \centering
  \includegraphics[width=0.76\textwidth]{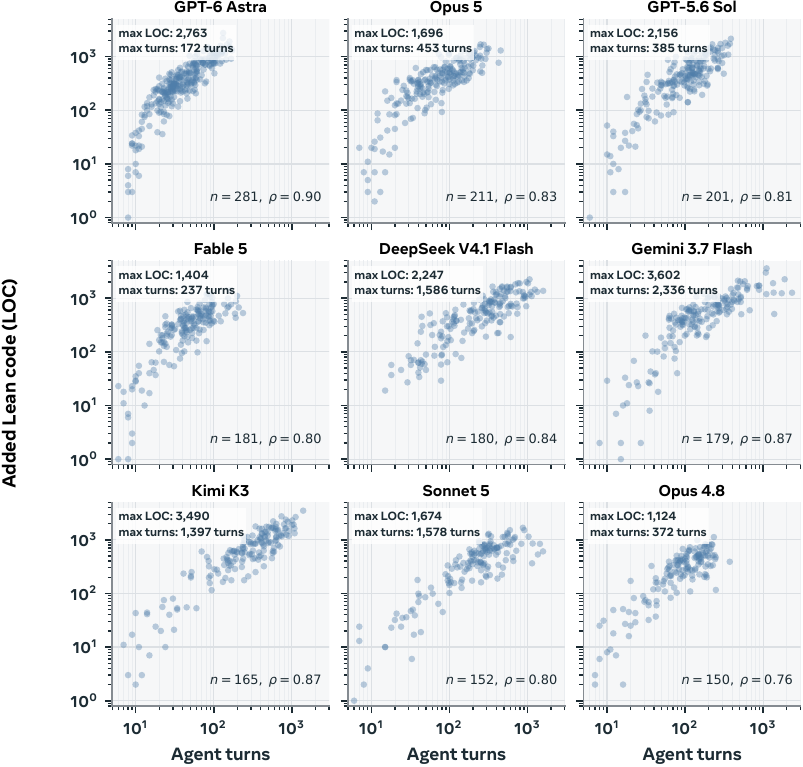}
  \caption{Added Lean code versus agent turns for every
  run, separated by model. Each panel reports the maximum added LOC and maximum
  number of turns. Both axes are logarithmic and shared across panels.}
  \label{fig:successful-loc-turns}
\end{figure}

\begin{figure}[H]
  \centering
  \includegraphics[width=0.76\textwidth]{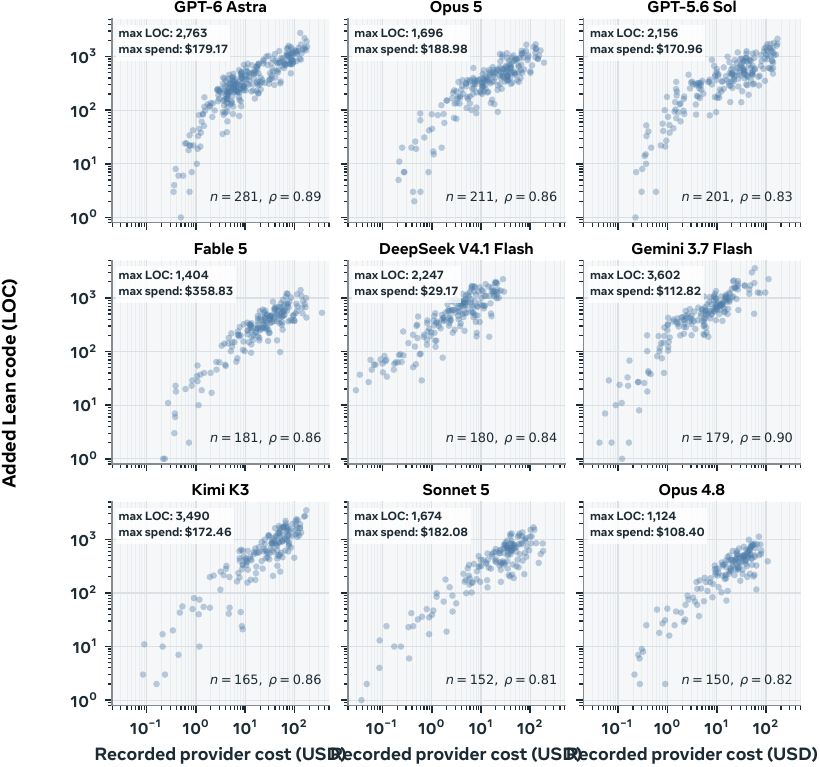}
  \caption{Added Lean code versus recorded provider spending for every run, separated by model. Each panel reports the maximum added
  LOC and maximum recorded spending. Both axes are logarithmic and shared
  across panels. The Fable maximum includes inseparable retry usage in its
  retained record.}
  \label{fig:successful-loc-cost}
\end{figure}

Finally, we broaden from Astra to every run and compare
raw added code with the resources used to produce it.
Figures~\ref{fig:successful-loc-turns} and~\ref{fig:successful-loc-cost} show
that generated code is strongly associated with both interaction length and
spending: within-model Spearman correlations range from 0.76 to 0.90 for turns
and from 0.81 to 0.90 for cost. Because models solve different task sets, these
plots characterize successful-run behavior rather than a controlled comparison.

\section{Run-to-Run Stability: Panel Construction and Detailed Results}
\label{app:stability-panel}

Repeating every benchmark item several times is unusually expensive in our
long-horizon setting. We therefore freeze a 72-problem panel before observing
any of the new repeatability runs. The eligible population is the set of problems that both GPT-5.6 Sol and Gemini~3.7 Flash have solved in our main set of runs.

Each problem receives one of four joint labels, with the first letter denoting
Sol and the second Gemini. Table~\ref{tab:stability-outcome-quotas} gives their
population counts and panel quotas. Quotas are obtained by multiplying the
population proportions by 72 and applying largest-remainder rounding.

\begin{table}[h]
  \centering
  \small
  \begin{tabular}{llrr}
    \toprule
    Label & Historical headline outcome & Full cohort & Panel \\
    \midrule
    SS & Both solved    & 173 & 37 \\
    FF & Neither solved & 134 & 28 \\
    SF & Sol only       &  28 &  6 \\
    FS & Gemini only    &   6 &  1 \\
    \bottomrule
  \end{tabular}
  \caption{Historical outcome strata used to construct the repeatability panel.}
  \label{tab:stability-outcome-quotas}
\end{table}

We selected a deterministic stratified panel of 72 tasks, balanced across all
24 textbooks, the historical outcomes in
Table~\ref{tab:stability-outcome-quotas}, and three structural-size measures:
workspace size, number of graded proof obligations, and target-statement size.
The task list and selection seed are released for reproducibility.

\paragraph{Repeated-run protocol and interpretation.}
For each model and panel item, we record three completed runs. Provider or
transport failures are replaced under the same replicate identifier. We report
the solve rate of each run, pass@3, and the number of items whose outcome agrees
in all three. These quantities describe the fixed panel and are not extrapolated
to full-benchmark pass@3.

\paragraph{Aggregate outcomes.}
Table~\ref{tab:stability-run-totals} gives the individual-run totals and their
union. Figure~\ref{fig:stability-budget} in the main text additionally shows
mean single-run coverage, its observed range, and pass@3 across resource
budgets.

\begin{table}[h]
  \centering
  \small
  \begin{tabular}{lrrrrr}
    \toprule
    Model & Run 1 & Run 2 & Run 3 & pass@3 & Unanimous \\
    \midrule
    GPT-5.6 Sol      & 43 & 44 & 43 & 45 & 69/72 \\
    Gemini~3.7 Flash & 38 & 36 & 38 & 39 & 69/72 \\
    \bottomrule
  \end{tabular}
  \caption{Verified outcomes across the three runs on the 72-problem panel.
  Pass@3 counts their union; unanimous counts problems with the same outcome in
  all three runs.}
  \label{tab:stability-run-totals}
\end{table}

\paragraph{Strategies across successful runs.}
We compare the three accepted proofs for every problem solved in all three runs,
using their target dependency closures and a manual comparison of the central
mathematical argument. GPT-5.6 Sol follows the same high-level route in 38/42
cases (90.5\%), and Gemini~3.7 Flash in 34/36 (94.4\%). For example, all runs
prove Boolean-quotient Sperner through Stanley's up/down operators, injectivity,
Hall matchings, and rank symmetry despite markedly different Lean developments.
Genuine alternatives also occur: for a spectral-endpoint theorem, runs alternate
between compact-spectrum/continuous-functional-calculus and
centered-operator/resolvent arguments. Stable success therefore usually reflects
convergence on the same mathematics, with substantial freedom in its Lean
realization.

\section{Source Provenance}

\input{tables/book_sources}

\section{Correction to a Source Formula}
\label{app:source-correction}

Proposition~10.3 of \emph{Learning with Submodular Functions} contains a
typographical omission in its $\ell_1$ characterization of submodular
minimization. With the source's notation
\[
  s_-(V)=\sum_{k\in V}\min\{s_k,0\},
  \qquad
  \lVert s\rVert_1=\sum_{k\in V}|s_k|,
\]
the printed proposition states
\[
  \min_{A\subseteq V}F(A)
  =\max_{s\in B(F)}s_-(V)
  =F(V)-\min_{s\in B(F)}\lVert s\rVert_1.
\]
The final equality is missing factors of $1/2$. Indeed, the scalar identity
$|x|=x-2\min\{x,0\}$ gives
\[
  \lVert s\rVert_1=s(V)-2s_-(V),
  \qquad
  s_-(V)=\tfrac12\bigl(s(V)-\lVert s\rVert_1\bigr).
\]
Since every $s\in B(F)$ satisfies $s(V)=F(V)$, the corrected formula is
\[
  \min_{A\subseteq V}F(A)
  =\tfrac12F(V)-\tfrac12\min_{s\in B(F)}\lVert s\rVert_1.
\]

The audited Lean target uses this corrected statement. The accepted proof makes
the normalization explicit through the helper lemma
\begin{quote}\small\ttfamily
l1Norm s = modularSetFunction s Finset.univ\par
\hspace*{2em}- 2 * modularSetFunction (negativePart s) Finset.univ,
\end{quote}
then uses the defining base-polyhedron equality $s(V)=F(V)$ to derive the two
$1/2$ factors. Thus the benchmark corrects a source typo rather than weakening
or otherwise changing the intended mathematical result.

\section{Agent Behavior and Ablation Details}
\label{app:ablation-details}

\subsection{Informal-Source Ablation}
\label{app:source-cases}

\paragraph{Experimental design.}
We test the marginal value of informal mathematics once the formal task has
already been fixed. In the full-source condition, the agent receives the Lean
workspace together with the textbook material in \texttt{book/} and an explicit
instruction to consult it. In the formal-workspace-only condition, the agent
receives the same expert-written target statement and local definitions, but all source-specific prompt text are
removed.

\paragraph{Paired outcomes and source use.}
Across the original 341-task source-enabled campaigns, Opus accesses the
textbook in 323/341 selected traces, while Gemini's attempt cascade
contains source-use evidence for 330/341  problems.  For Opus, the paired outcomes are 198 successes in both conditions, 6 with-source-only successes, 2
without-source-only successes, and 135 failures in both conditions. Gemini has 126 successes in both
conditions, 10 with-source-only successes, 7 without-source-only successes, and
198 failures in both conditions.

Among tasks solved in both conditions, the source does not consistently shorten
the resulting Lean development. For Opus, the median with-source/without-source
LOC ratio is 1.00: the source-conditioned version is shorter in 99/198 pairs,
longer in 96, and equal in 3. For Gemini, the median ratio is 1.04: the
source-conditioned version is shorter in 55/126 pairs, longer in 68, and equal
in 3. These comparisons concern generated code length rather than mathematical
or software quality.

\paragraph{Illustrative cases.}
Opus~5 solved each of the following four problems only when given the textbook.
In every case, its trace shows that it read a relevant passage and used the
mathematical structure of that passage in the successful formalization.

\begin{itemize}[leftmargin=*]
  \item \textbf{Nagell--Lutz theorem.} The source supplied the sequence of local
  valuation and formal-group lemmas needed to prove integrality and the final
  discriminant divisibility statement. The no-source run developed much of the
  same machinery but did not connect it into the required torsion argument.

  \item \textbf{Rouch\'e's theorem.} Opus~5 followed the source's decomposition
  through the argument principle and a straight-line homotopy. The no-source run
  constructed the zero-counting machinery but stopped before proving its final
  connection to winding number.

  \item \textbf{Commutativity of Dunkl operators.} The successful run used the
  source's reduction through a commutator identity and the Jacobi identity,
  avoiding the classification argument pursued by the no-source run.

  \item \textbf{Universal splitting.} The source gave a three-part roadmap:
  representability by polynomial equations, formal smoothness by lifting matrix
  units, and faithful flatness through splitting over algebraically closed
  fibres. The successful run implemented all three; the no-source run stopped at
  the final faithful-flatness step.
\end{itemize}

Opus~5 also solved two problems only without the source:

\begin{itemize}[leftmargin=*]
  \item \textbf{Grading-element local finiteness.} The source-enabled run followed
  the textbook's quotient-module argument but stalled on two technical steps. The
  no-source run instead used local-finiteness saturations and generalized
  eigenspaces, which translated more directly into Lean.

  \item \textbf{Soft-thresholded SVD.} The source-enabled run followed a
  stationary-point argument whose relaxation lost an essential rank constraint.
  The no-source run instead orthogonalized the factors and completed the proof
  using scalar soft thresholding, convexity, and a fractional greedy bound.
\end{itemize}

\subsection{Reasoning-Effort Sensitivity}
\label{app:reasoning-effort}

Reasoning-effort controls allocate additional test-time computation within a
model, but their labels are provider-specific and should not be compared as
matched compute across model families. To measure within-model sensitivity, we
rerun four models below their headline settings: GPT-5.6 Sol and Gemini~3.7
Flash at \texttt{medium}, and Opus~5 and Fable~5 with adaptive thinking at
\texttt{high}.

The effect is sharply model dependent. Opus~5 drops only from 211/341 (61.9\%)
to 204/341 (59.8\%), a decrease of 2.1 percentage points, while Fable~5 remains
unchanged at 181/341 (53.1\%). In contrast, Gemini falls from 179/341 (52.5\%)
to 136/341 (39.9\%), a decrease of 12.6 percentage points, and GPT-5.6 Sol falls
from 201/341 (58.9\%) to 114/341 (33.4\%), a decrease of 25.5 percentage points.
Lower reasoning effort therefore has little observed effect on Fable and only a
modest effect on Opus, but causes a substantial loss of verified coverage for
Gemini and Sol.

\subsection{One-Shot and Tool-Ablation Details}

GPT-5.6 Sol's six one-shot successes are finite-dimensional norm equivalence,
Riesz representation, a convex-hull support identity, Jacobson density, the
Lasso optimality conditions, and a Stacks open-refinement lemma. Gemini's three
are universal representation by an isotropic quadratic form, the nonpositive
curvature inequality for affine buildings, and existence of a projective-module
surjection. The two success sets are disjoint, but all nine problems are solved
by every model in the agentic leaderboard, including GPT-6 Astra, DeepSeek
V4.1 Flash, and Kimi K3. Their corresponding full agentic runs require far fewer
interactions than a typical success on the single-hole cohort: a median of 19
versus 102 turns for Sol and 17 versus 137.5 for Gemini. The one-shot comparison
measures the complete one-shot and agentic systems rather than the effect of any
single tool.

The REPL, Loogle, and build endpoints are heavily used in the three baseline
runs underlying pass@3. Across 216 task--run observations, GPT makes 8{,}715
REPL calls, 18{,}459 Loogle calls, and 5{,}159 build-service calls; Gemini makes
33{,}209, 12{,}416, and 19{,}058, respectively. Every observation invokes build,
and 214/216 GPT observations and 215/216 Gemini observations use both REPL and
build. Tool use is therefore a central part of the baseline behavior rather
than a rarely exercised option.

\begin{table}[H]
  \centering
  \small
  \begin{tabular}{lrr}
    \toprule
    Tool group & GPT-5.6 Sol & Gemini~3.7 Flash \\
    \midrule
    REPL        & 36 [17, 58]       (99.1\%) & 106.5 [40.5, 217.8] (99.5\%) \\
    Loogle      & 62 [24, 125.3]    (97.7\%) & 46.5 [18.8, 82]    (97.2\%) \\
    Build       & 20 [9, 35.3]      (100\%)  & 69 [11.8, 138.5]   (100\%) \\
    \bottomrule
  \end{tabular}
  \caption{Baseline tool calls per task--run, reported as median
  [first quartile, third quartile] across the 216 pass@3 observations for each
  model. Parentheses give the percentage of observations using the tool group.}
  \label{tab:baseline-tool-use}
\end{table}

Because Gemini trajectories are longer, raw calls per run should not be read as
cross-model efficiency. Per 100 turns, GPT makes 23.7 REPL, 50.1 Loogle, and
14.0 build-service calls, compared with 30.7, 11.5, and 17.6 for Gemini. Failed
runs use more tools because successful runs terminate early while failures
commonly run to budget.

Filesystem search provides a fallback when specialized search is removed:
median filesystem-read calls rise from roughly 7 per GPT full-tool run to 55
without Loogle and 85 with only file operations; for Gemini, they rise from 20
to 40 and 52, respectively. In the no-build condition, GPT reconstructs
\texttt{lake build} through the REPL on 26/72 problems, illustrating why these
treatments concern endpoints rather than abstract capabilities. Among the 34
problems solved by Gemini both with full tools and without the REPL, median
task-level spending increases by approximately $1.84\times$ without the REPL.
GPT shows no comparable increase, suggesting that the REPL primarily improves
Gemini's efficiency rather than its aggregate success rate.

\subsection{Alternative Agent Scaffold}

Claude Code uses a hybrid tool stack rather than a wholly separate proving
environment. Across its selected traces, 60.6\% of tool calls go to the
benchmark-provided MCP services for the Lean REPL, Loogle, build, and report
submission; the remaining 39.4\% use Claude Code's native \texttt{Read},
\texttt{Glob}, \texttt{Grep}, \texttt{Edit}, and \texttt{Write} operations. It
invokes the REPL on 312/341 problems, Loogle on 301, and build on 317.

In 18 of the 20 Claude-only successes, both configurations follow the same
mathematical roadmap, but the common loop stops with only one or two target
holes open. Claude Code's accepted developments have median 1{,}402 added lines
in these cases, compared with 415 in the common loop's partial submissions; it
performs more persistent editing while making fewer REPL and Loogle calls. For
example, on B\'ezout's theorem the common loop makes 55 Loogle calls and adds
442 lines but leaves the homological core unfinished; Claude Code uses 38
Loogle calls and adds 1{,}740 lines to complete the Hilbert--Koszul and
local-intersection-algebra development.
Claude Code has
lower recorded spending on all 20 problems (median \$56 versus \$120) and uses
fewer turns on 17/20 (median 235 versus 305).

\subsection{Persistence Intervention}
\label{app:persistence}

Early in our experiments, agents could end a run by reporting a problem
\texttt{unsolved} after using only a small fraction of their budget. We therefore
introduced the 10\% continuation rule described in
Appendix~\ref{app:evaluation-harness}. In the canonical leaderboard exports,
more than one valid \texttt{submit\_report} call identifies a selected trajectory
whose first surrender was rejected. Table~\ref{tab:persistence} reports how often
this occurred and how often the additional work eventually produced a verified
proof.

\begin{table}[H]
  \centering
  \small
  \setlength{\tabcolsep}{6pt}
  \renewcommand{\arraystretch}{1.08}
  \begin{tabular}{lrrr}
    \toprule
    Model & Forced to continue & Eventually passed & Conversion \\
    \midrule
    GPT-6 Astra & 13 & 12 & 92.3\% \\
    Opus~5 & 3 & 0 & 0\% \\
    GPT-5.6 Sol & 88 & 17 & 19.3\% \\
    Fable~5 & 2 & 0 & 0\% \\
    DeepSeek V4.1 Flash & 198 & 39 & 19.7\% \\
    Gemini~3.7 Flash & 199 & 41 & 20.6\% \\
    Kimi K3 & 84 & 3 & 3.6\% \\
    Sonnet~5 & 114 & 6 & 5.3\% \\
    Opus~4.8 & 134 & 0 & 0\% \\
    \bottomrule
  \end{tabular}
  \caption{Effect of requiring agents to continue after an early
  \texttt{unsolved} report in the selected headline trajectories. Conversion is
  the fraction of forced continuations that eventually pass clean-room
  verification.}
  \label{tab:persistence}
\end{table}

Continued work recovers proofs for Astra, Gemini, DeepSeek, Sol, Sonnet, and
Kimi, but none for Opus~5, Opus~4.8, or Fable. The intervention
is therefore most useful for models that tend to stop before exhausting their
productive search, and offers little benefit when an \texttt{unsolved} report
already reflects a stable failure.

\paragraph{Impact on headline scores.}
The canonical leaderboard exports record the number of valid
\texttt{submit\_report} calls made in each selected trajectory. A verified run
with more than one call necessarily submitted \texttt{unsolved} first: a
\texttt{solved} report terminates immediately. We therefore mechanically
truncate each trajectory at its first surrender to obtain
Table~\ref{tab:early-surrender-score}. Without enforced continuation, Astra
would fall from 281 to 269, DeepSeek from 180 to 141, Gemini from 179 to 138,
and GPT-5.6 Sol from 201 to 184. The gains show that sustained iteration on compiler and tool feedback is
itself an important form of test-time scaling.

\begin{table}[H]
  \centering
  \small
  \begin{tabular}{lrrr}
    \toprule
    Model & Final score & Later recoveries & Stop-at-surrender score \\
    \midrule
    GPT-6 Astra         & 281 & 12 & 269 \\
    Opus~5              & 211 & 0  & 211 \\
    GPT-5.6 Sol         & 201 & 17 & 184 \\
    Fable~5             & 181 & 0  & 181 \\
    DeepSeek V4.1 Flash & 180 & 39 & 141 \\
    Gemini~3.7 Flash    & 179 & 41 & 138 \\
    Kimi K3             & 165 & 3  & 162 \\
    Sonnet~5            & 152 & 6  & 146 \\
    Opus~4.8            & 150 & 0  & 150 \\
    \bottomrule
  \end{tabular}
  \caption{Headline verified coverage with the continuation intervention and
  after counterfactually terminating each selected trajectory at its first
  valid \texttt{unsolved} report.}
  \label{tab:early-surrender-score}
\end{table}

\begin{table}[H]
  \centering
  \small
  \begin{tabular}{lrr}
    \toprule
    Model & Budget consumed & Model gave up \\
    \midrule
    GPT-6 Astra         & 21 & 39 \\
    GPT-5.6 Sol         & 10 & 130 \\
    Sonnet~5            & 4  & 185 \\
    Kimi K3             & 3  & 173 \\
    Opus~5              & 2  & 128 \\
    Gemini~3.7 Flash    & 2  & 160 \\
    Fable~5             & 0  & 160 \\
    DeepSeek V4.1 Flash & 0  & 161 \\
    Opus~4.8            & 0  & 191 \\
    \midrule
    All models          & 42 & 1{,}327 \\
    \bottomrule
  \end{tabular}
  \caption{Termination reasons among unsuccessful selected leaderboard
  trajectories.}
  \label{tab:termination-reasons}
\end{table}

\subsection{Self-Report Accuracy}
\label{app:self-report}

Agents generally assess their own success accurately, but their reports are not
perfect substitutes for independent grading. Table~\ref{tab:self-report} compares
each recorded final \texttt{solved}/\texttt{unsolved} report with the clean-room
verdict. Of 2{,}969 recorded reports, 2{,}959 agree with the grader. All 10
disagreements are false claims of success; no model reports \texttt{unsolved} for
a submission that Comparator accepts. A further 100 rows have no recorded final
report and are shown separately rather than counted as correct or incorrect.

\begin{table}[H]
  \centering
  \small
  \setlength{\tabcolsep}{5pt}
  \renewcommand{\arraystretch}{1.08}
  \begin{tabular}{lrrrr}
    \toprule
    Model & Correct/recorded & False \texttt{solved} & No report & Accuracy \\
    \midrule
    GPT-6 Astra & 339/341 & 2 & 0 & 99.41\% \\
    Opus~5 & 341/341 & 0 & 0 & 100.00\% \\
    Opus~4.8 & 341/341 & 0 & 0 & 100.00\% \\
    GPT-5.6 Sol & 340/341 & 1 & 0 & 99.71\% \\
    Gemini~3.7 Flash & 336/340 & 4 & 1 & 98.82\% \\
    Sonnet~5 & 340/341 & 1 & 0 & 99.71\% \\
    Fable~5 & 243/243 & 0 & 0 & 100.00\% \\
    DeepSeek V4.1 Flash & 340/341 & 1 & 0 & 99.71\% \\
    Kimi K3 & 339/340 & 1 & 1 & 99.71\% \\
    \midrule
    All models & 2{,}959/2{,}969 & 10 & 100 & 99.66\% \\
    \bottomrule
  \end{tabular}
  \caption{Agreement between recorded agent self-reports and clean-room
  grading. Accuracy is conditional on a recorded final report.}
  \label{tab:self-report}
\end{table}

\end{document}

%% file: packages.tex
\usepackage{tikz}
\usepackage{pgfplots}
\pgfplotsset{compat=1.17}

\usepackage{array}
\usepackage{enumitem}
\setitemize{itemsep=0pt,parsep=0pt,topsep=0pt}
\setenumerate{itemsep=0pt,parsep=0pt,topsep=0pt}

\usepackage{float}
\usepackage{wrapfig}
\usepackage{algorithm}
\usepackage{algorithmicx}
\usepackage{algpseudocode}
\usepackage[normalem]{ulem}

%% file: tables/book_results.tex
\begin{table}[H]
  \caption{Per-book verified solve counts on the 341-task benchmark. Model
  counts use the headline condition at the highest evaluated effort.
  \textbf{Astra}: GPT-6 Astra; \textbf{O5}: Opus 5;
  \textbf{Sol}: GPT-5.6 Sol;
  \textbf{F5}: Fable 5; \textbf{DS}: DeepSeek V4.1 Flash;
  \textbf{G3.7F}: Gemini 3.7 Flash; \textbf{K3}: Kimi K3;
  \textbf{S5}: Sonnet 5; \textbf{O4.8}: Opus 4.8. \textbf{Any} counts tasks
  solved by at least one model. Row-wise best
  model counts are bold.}
  \label{tab:book-results}
  \centering
  \scriptsize
  \setlength{\tabcolsep}{1.8pt}
  \renewcommand{\arraystretch}{1.08}
  \resizebox{\textwidth}{!}{%
  \begin{tabular}{lrrrrrrrrrrr}
    \toprule
    Book & Tasks & Astra & O5 & Sol & F5 & DS & G3.7F & K3 & S5 & O4.8 & Any \\
    \midrule
    \emph{Algebraic Combinatorics} & 2 & \textbf{2} & \textbf{2} & \textbf{2} & \textbf{2} & 1 & 1 & \textbf{2} & 1 & \textbf{2} & 2 \\
    \emph{Algebraic Geometry} & 8 & \textbf{6} & 1 & 1 & 1 & 1 & 1 & 1 & 0 & 0 & 6 \\
    \emph{Algebraic Topology} & 16 & \textbf{10} & 5 & 4 & 5 & 3 & 3 & 3 & 3 & 2 & 10 \\
    \emph{Arithmetic Geometry} & 26 & \textbf{20} & 16 & 16 & 13 & 13 & 14 & 11 & 10 & 12 & 20 \\
    \emph{Boolean Functions} & 2 & \textbf{2} & \textbf{2} & \textbf{2} & \textbf{2} & \textbf{2} & 1 & 1 & 0 & 1 & 2 \\
    \emph{Buildings} & 11 & \textbf{11} & 9 & 10 & 7 & 7 & 9 & 7 & 5 & 6 & 11 \\
    \emph{Cherednik Algebras} & 15 & \textbf{12} & 6 & 4 & 3 & 3 & 2 & 4 & 1 & 1 & 12 \\
    \emph{Complex Variables} & 12 & \textbf{12} & \textbf{12} & 10 & 9 & 9 & 8 & 9 & 9 & 8 & 12 \\
    \emph{Differential Analysis} & 18 & \textbf{17} & 14 & 13 & 12 & 14 & 10 & 14 & 12 & 12 & 17 \\
    \emph{Elliptic Curves} & 27 & \textbf{11} & 7 & 7 & 5 & 5 & 7 & 5 & 4 & 4 & 12 \\
    \emph{Graph/Additive Comb.} & 9 & \textbf{6} & 4 & 4 & 4 & 4 & 4 & 2 & 4 & 4 & 6 \\
    \emph{High-Dimensional Statistics} & 25 & \textbf{25} & \textbf{25} & \textbf{25} & 24 & 24 & 22 & 22 & 22 & 21 & 25 \\
    \emph{Lie Groups and Algebras} & 28 & \textbf{18} & 11 & 12 & 10 & 11 & 10 & 9 & 10 & 8 & 18 \\
    \emph{Low-Rank Models} & 5 & \textbf{5} & 4 & 4 & 4 & 3 & 4 & 3 & 3 & 3 & 5 \\
    \emph{Matrix Concentration} & 3 & \textbf{3} & 0 & 0 & 0 & 0 & 0 & 0 & 0 & 0 & 3 \\
    \emph{Noncommutative Algebra} & 22 & \textbf{17} & 14 & 14 & 11 & 13 & 12 & 10 & 10 & 9 & 17 \\
    \emph{Number Theory} & 24 & \textbf{18} & 8 & 8 & 5 & 5 & 5 & 5 & 3 & 3 & 18 \\
    \emph{PDEs} & 8 & \textbf{8} & \textbf{8} & 7 & 7 & 6 & 7 & 7 & 6 & 6 & 8 \\
    \emph{Probability} & 4 & \textbf{4} & \textbf{4} & 3 & \textbf{4} & \textbf{4} & \textbf{4} & \textbf{4} & 3 & 2 & 4 \\
    \emph{Sparsity Penalties} & 8 & \textbf{8} & \textbf{8} & \textbf{8} & \textbf{8} & \textbf{8} & \textbf{8} & 7 & 7 & 7 & 8 \\
    \emph{Stacks Project} & 32 & \textbf{31} & 28 & 24 & 24 & 22 & 27 & 20 & 20 & 20 & 31 \\
    \emph{Submodular Functions} & 15 & \textbf{15} & \textbf{15} & 14 & 13 & 14 & 13 & 12 & 11 & 13 & 15 \\
    \emph{Tensor Categories} & 13 & \textbf{13} & 4 & 5 & 4 & 4 & 3 & 3 & 4 & 2 & 13 \\
    \emph{Theory of Computation} & 8 & \textbf{7} & 4 & 4 & 4 & 4 & 4 & 4 & 4 & 4 & 7 \\
    \midrule
    \textbf{Total (24 books)} & \textbf{341} & \textbf{281} & \textbf{211} & \textbf{201} & \textbf{181} & \textbf{180} & \textbf{179} & \textbf{165} & \textbf{152} & \textbf{150} & \textbf{282} \\
    \bottomrule
  \end{tabular}%
  }
\end{table}

%% file: tables/book_sources.tex
\begin{table}[H]
  \caption{Provenance of the 24 mathematical sources used in AutoformBench. Each
  link points to the canonical public course page, project page, or arXiv
  record from which the benchmark source was obtained\vspace{5pt} }
  \label{tab:source-provenance}
  \centering
  \scriptsize
  \setlength{\tabcolsep}{3.5pt}
  \renewcommand{\arraystretch}{1.04}
  \begin{tabular}{p{0.42\textwidth}p{0.40\textwidth}l}
    \toprule
    Source & Author(s) & URL \\
    \midrule
    \emph{Topics in Algebraic Combinatorics} & Richard P. Stanley & \href{https://ocw.mit.edu/courses/18-318-topics-in-algebraic-combinatorics-spring-2006/}{MIT OCW} \\
    \emph{Algebraic Geometry I} & Roman Bezrukavnikov et al. & \href{https://ocw.mit.edu/courses/18-725-algebraic-geometry-fall-2015/}{MIT OCW} \\
    \emph{Lectures on Algebraic Topology} & Haynes Miller; notes by Sanath Devalapurkar & \href{https://ocw.mit.edu/courses/18-905-algebraic-topology-i-fall-2016/}{MIT OCW} \\
    \emph{Introduction to Arithmetic Geometry} & Andrew V. Sutherland & \href{https://ocw.mit.edu/courses/18-782-introduction-to-arithmetic-geometry-fall-2013/}{MIT OCW} \\
    \emph{Analysis of Boolean Functions} & Dor Minzer & \href{https://ocw.mit.edu/courses/18-218-topics-in-combinatorics-analysis-of-boolean-functions-spring-2021/}{MIT OCW} \\
    \emph{Buildings} & Paul Garrett & \href{https://www-users.cse.umn.edu/~garrett/m/buildings/book.pdf}{author PDF} \\
    \emph{Lecture Notes on Cherednik Algebras} & Pavel Etingof and Xiaoguang Ma & \href{https://arxiv.org/abs/1001.0432}{arXiv:1001.0432} \\
    \emph{Functions of a Complex Variable} & Sigurdur Helgason & \href{https://ocw.mit.edu/courses/18-112-functions-of-a-complex-variable-fall-2008/}{MIT OCW} \\
    \emph{Differential Analysis} & Richard B. Melrose & \href{https://ocw.mit.edu/courses/18-155-differential-analysis-fall-2004/}{MIT OCW} \\
    \emph{Elliptic Curves} & Andrew V. Sutherland & \href{https://ocw.mit.edu/courses/18-783-elliptic-curves-spring-2021/}{MIT OCW} \\
    \emph{Graph Theory and Additive Combinatorics} & Yufei Zhao & \href{https://ocw.mit.edu/courses/18-225-graph-theory-and-additive-combinatorics-fall-2023/}{MIT OCW} \\
    \emph{High-Dimensional Statistics} & Philippe Rigollet and Jan-Christian H\"utter & \href{https://ocw.mit.edu/courses/18-s997-high-dimensional-statistics-spring-2015/}{MIT OCW} \\
    \emph{Lie Groups and Lie Algebras} & Pavel Etingof & \href{https://arxiv.org/abs/2201.09397}{arXiv:2201.09397} \\
    \emph{Generalized Low Rank Models} & Madeleine Udell, Corinne Horn, Reza Zadeh, and Stephen Boyd & \href{https://arxiv.org/abs/1410.0342}{arXiv:1410.0342} \\
    \emph{An Introduction to Matrix Concentration Inequalities} & Joel A. Tropp & \href{https://arxiv.org/abs/1501.01571}{arXiv:1501.01571} \\
    \emph{Noncommutative Algebra} & Roman Bezrukavnikov & \href{https://ocw.mit.edu/courses/18-706-noncommutative-algebra-spring-2023/}{MIT OCW} \\
    \emph{Number Theory I} & Andrew V. Sutherland & \href{https://ocw.mit.edu/courses/18-785-number-theory-i-fall-2021/}{MIT OCW} \\
    \emph{Introduction to Partial Differential Equations} & Jared Speck & \href{https://ocw.mit.edu/courses/18-152-introduction-to-partial-differential-equations-fall-2011/}{MIT OCW} \\
    \emph{Theory of Probability} & Scott Sheffield & \href{https://ocw.mit.edu/courses/18-175-theory-of-probability-spring-2014/}{MIT OCW} \\
    \emph{Optimization with Sparsity-Inducing Penalties} & Francis Bach, Rodolphe Jenatton, Julien Mairal, and Guillaume Obozinski & \href{https://arxiv.org/abs/1108.0775}{arXiv:1108.0775} \\
    \emph{The Stacks Project} & The Stacks Project Authors & \href{https://stacks.math.columbia.edu/}{project site} \\
    \emph{Learning with Submodular Functions} & Francis Bach & \href{https://arxiv.org/abs/1111.6453}{arXiv:1111.6453} \\
    \emph{Tensor Categories} & Pavel Etingof, Shlomo Gelaki, Dmitri Nikshych, and Victor Ostrik & \href{https://ocw.mit.edu/courses/18-769-topics-in-lie-theory-tensor-categories-spring-2009/}{MIT OCW} \\
    \emph{Introduction to the Theory of Computation} & Michael Sipser & \href{https://ocw.mit.edu/courses/18-404j-theory-of-computation-fall-2020/}{MIT OCW} \\
    \bottomrule
  \end{tabular}
\end{table}